\documentclass[10pt,journal,compsoc]{IEEEtran}

\usepackage{caption}
\usepackage{times}
\usepackage{epsfig}
\usepackage{graphicx}
\usepackage{amsmath}
\usepackage{amssymb}

\usepackage{bbding}      

\usepackage{multirow}
\usepackage{subfigure}
\usepackage{ulem}
\usepackage{color} 
\usepackage{enumerate}

\usepackage[backref]{hyperref}
\usepackage{makecell}

\ifCLASSOPTIONcompsoc
  \usepackage[nocompress]{cite}
\else
  \usepackage{cite}
\fi
\ifCLASSINFOpdf
\else
\fi
\begin{document}
%
\title{Identity-Guided Collaborative Learning for Cloth-Changing Person Reidentification}
%
%
%
%

\author{Zan Gao*, ~\IEEEmembership{Member,~IEEE}, Shengxun Wei*, Weili Guan, Lei Zhu, ~\IEEEmembership{Senior Member,~IEEE,}\\ Meng Wang, ~\IEEEmembership{Fellow,~IEEE,} and Shengyong Chen, ~\IEEEmembership{Senior Member,~IEEE,} IET Fellow

\IEEEcompsocitemizethanks{\IEEEcompsocthanksitem Z. Gao and S.X Wei are with the Shandong Artificial Intelligence Institute, Qilu University of Technology (Shandong Academy of Sciences), Jinan, 250014, P.R China. (e-mail: zangaonsh4522@gmail.com, wei\_shengxun@163.com) 

\IEEEcompsocthanksitem S.Y Chen and Z. Gao are with the Key Laboratory of Computer Vision and System, Ministry of Education, Tianjin University of Technology, Tianjin, 300384, China. (e-mail: sy@ieee.org, zangaonsh4522@gmail.com)

\IEEEcompsocthanksitem W.L Guan is with the Faculty of Information Technology, Monash University Clayton Campus, Australia. (e-mail: honeyguan@gmail.com).

\IEEEcompsocthanksitem L. Zhu is with the School of Electronic and Information Engineering, Tongji University, 200092, Shanghai, China (e-mail: leizhu0608@gmail.com).

\IEEEcompsocthanksitem M. Wang is with the school of Computer Science and Information Engineering, Hefei University of Technology, Hefei, 230009, P.R China. (e-mail:eric.mengwang@gmail.com) 

\IEEEcompsocthanksitem Corresponding author: Shengxun Wei and Zan Gao
\protect\\
}
}

\markboth{IEEE TRANSACTIONS ON PATTERN ANALYSIS AND MACHINE INTELLIGENCE, 2023}%
{Shell \MakeLowercase{\textit{et al.}}: Bare Demo of IEEEtran.cls for Computer Society Journals}
%



\IEEEtitleabstractindextext{%
\begin{abstract} Cloth-changing person reidentification (ReID) is a newly emerging research topic aimed at addressing the issues of large feature variations due to cloth-changing and pedestrian view/pose changes. Although significant progress has been achieved by introducing extra information (e.g., human contour sketching information, human body keypoints, and 3D human information), cloth-changing person ReID remains challenging because pedestrian appearance representations can change at any time. Moreover, human semantic information and pedestrian identity information are not fully explored. To solve these issues, we propose a novel identity-guided collaborative learning scheme (IGCL) for cloth-changing person ReID, where the human semantic is effectively utilized and the identity is unchangeable to guide collaborative learning. First, we design a novel clothing attention degradation stream to reasonably reduce the interference caused by clothing information where clothing attention and mid-level collaborative learning are employed. Second, we propose a human semantic attention and body jigsaw stream to highlight the human semantic information and simulate different poses of the same identity. In this way, the extraction features not only focus on human semantic information that is unrelated to the background but are also suitable for pedestrian pose variations. Moreover, a pedestrian identity enhancement stream is  proposed to enhance the identity importance and extract more favorable identity robust features. Most importantly, all these streams are jointly explored in an end-to-end unified framework, and the identity is utilized to guide the optimization. Extensive experiments on six public clothing person ReID datasets (LaST, LTCC, PRCC, NKUP, Celeb-reID-light, and VC-Clothes) demonstrate the superiority of the IGCL method. It outperforms existing methods on multiple datasets, and the extracted features have stronger representation and discrimination ability and are weakly correlated with clothing.

\end{abstract}

}

\maketitle

\IEEEdisplaynontitleabstractindextext

%
\IEEEpeerreviewmaketitle

\IEEEraisesectionheading{\section{Introduction}\label{sec:introduction}}

%
%
%
%

\IEEEPARstart{P}{erson} reidentification (ReID) is an active research topic in computer vision and machine learning. Its aim is to match pedestrians with the same identity across different cameras. In the last decade, this task has achieved significant progress. However most of the literature is based on the assumption that a person's clothes will not change, and the visual appearances of pedestrians are required to have the same clothes. In real conditions, cloth-changing often occurs when the surveillance acquisition period is extended. If the existing person ReID approaches are directly applied in this case, their performances substantially deteriorate and often fail. Some researchers have paid more attention to investigating the cloth-changing person ReID task \cite{gao2022multigranular, yang2021clothing}, i.e., searching for the same pedestrian with a piece of clothing in other camera views when given only a probe image with another piece of clothing.

\begin{figure}[t]
\begin{center}
\includegraphics[width=3.5in,height = 2.5in]{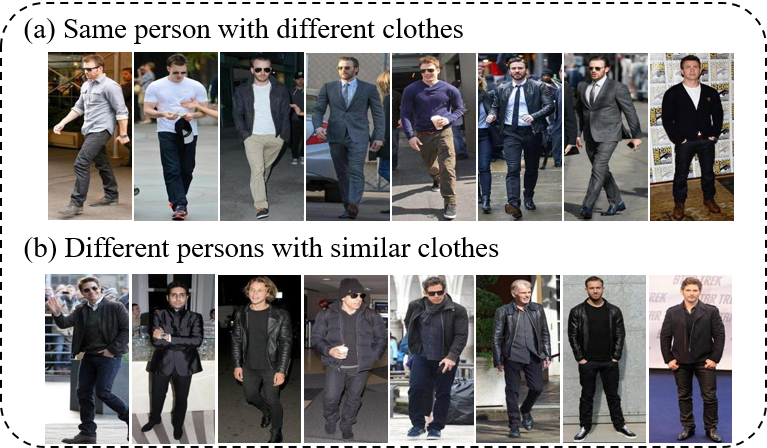}
\caption{Examples of cloth-changing person ReID images. The color appearance of the pedestrian samples in the cloth-changing scene exhibits large intraclass changes and small interclass changes.} \label{Example_Figure}
\end{center}
\vspace{-2.5em}
\end{figure}

A few researchers \cite{Zhang23Specialized, gao2022multigranular, huang2020beyond,  qian2020long, zheng2019joint, yu2020cocas, yang2021clothing} have made useful attempts in the cloth-changing person ReID task. Previously, to promote the development of the cloth-changing person ReID task, different datasets, such as LTCC \cite{qian2020long}, PRCC \cite{yang2021clothing}, Celeb-reID \cite{huang2020beyond}, and NKUP \cite{wang2020benchmark}, were built. Recently, a few researchers have proposed novel cloth-changing person ReID approaches. For example, Yang et al. \cite{yang2021clothing} proposed the SPT+ASE module, where the human contour sketch was employed to decouple the color information of an image and introduced a spatial polar transformation (SPT) layer into a deep neural network to transform the contour sketch. Hong et al. \cite{Hong2021FSAM} proposed a novel fine-grained shape-appearance mutual learning framework (FSAM) to interactively learn between low-level features and high-level features to transfer knowledge from the shape stream to the appearance stream. This approach allows the appearance stream to be independently deployed without needing additional computation for mask estimation.

In the cloth-changing person ReID task, there are two main challenges: I) 
a pedestrian with different pieces of clothing
 and II) changes in pedestrians' views/poses. Despite the performance gains achieved by previous research to address these issues, several limitations remain: 1) \textbf{Insufficient representations}. Since the human appearance exhibits large variations with different clothes, for example, Figure \ref{Example_Figure} (a) displays the differences for the same person with a different piece of clothing, and Figure \ref{Example_Figure}(b) shows the similarities among different persons with similar clothes. Thus, it is very difficult for existing approaches \cite{Gu2022Clothes, Hong2021FSAM} to extract discriminative and robust feature representations. Therefore, reducing the influence of cloth-changing to significantly improve the robustness and discrimination of their visual representations is a problem that urgently needs to be addressed. 2) \textbf{Underdeveloped semantic information}. Although human semantic information is  used in many existing ReID methods, they mainly focus on body shape or contour sketches, and human semantic information is not effectively explored (in other words, the meanings of different parts of the human body are often ignored). Thus, how to adequately take advantage of human semantic information is still underexplored. 3) \textbf{Underemphasizing pedestrian identity}. Most existing methods \cite{gao2022multigranular, Gu2022Clothes} do not focus on pedestrian identity and often learn identity classification features implicitly from original features without emphasizing the parts strongly related to identity according to pedestrian characteristics. Thus, explicitly guiding the model to learn identity classification features and emphasizing identity importance is worth exploring.

To address the abovementioned problems, we design a novel IGCL framework to exploit robust and informative pedestrian representations for cloth-changing person ReID. To address the problem (1), we propose a unified and end-to-end collaborative learning network architecture where different streams are jointly explored. Moreover, we introduce the clothing-attention map activation module to reduce the interference caused by clothing information. In this way, more robust and discriminative feature representations can be extracted. To solve the problem (2), we propose human semantic attention and a body jigsaw module to highlight the human semantic information and enrich the sample distribution of different poses of the same identity. In addition, to address the problem (3), we design a pedestrian identity enhancement module to enhance identity importance. Moreover, in each module, the pedestrian identity is kept unchanged. Experimental results for six public, cloth-changing person ReID datasets validate the superiority of our framework. The main contributions of this paper are as follows:
\begin{itemize}
\item We develop a novel scheme for cloth-changing person ReID that jointly integrates four different streams into an end-to-end unified framework. Moreover, the human semantic is effectively utilized, and the identity
is kept unchanged
to guide collaborative learning. In this way, the extracted feature is more robust, discriminative, and clothing-irrelevant.

\item We design a novel clothing attention degradation stream (CAD) to reduce the interference caused by clothing information, where the importance of the clothing area is weakened, and clothing attention and mid-level collaborative learning are employed. In addition, we propose human semantic attention and a body jigsaw stream (SAJ) to highlight human semantic information and simulate different poses of the same identity. We develop a pedestrian identity enhancement stream (PIE) to enhance identity importance, where only head and shoulder information is employed. Thus, more favorable identity robust features can be extracted to recognize pedestrian identity. Note that "mid-level collaborative learning" refers to distilling low-weight features of clothing regions at multiple scales.

\item We systematically and comprehensively evaluate the proposed IGCL on six public, cloth-changing person ReID datasets, including LaST, LTCC, PRCC, NKUP, Celeb-reID-light, and VC-Clothes. Experimental results show that IGCL obtains more robust and discriminative identity-related features, while effectively reducing clothing interference. IGCL has significant mAP and Rank-1 improvement on multiple datasets.
\end{itemize}

The remainder of the paper is organized as follows: Section II introduces related work, and Section III describes the proposed IGCL method. Section IV describes the experimental settings, results, and an analysis of the results. Section V presents details of the ablation study, and concluding remarks are presented in Section VI.

\section{Related Work}

Since person ReID has an important role in surveillance video analysis, many researchers have paid attention to this topic and have proposed many methods. According to a person's visual appearance, these methods can be roughly divided into clothing-consistent person ReID and cloth-changing person ReID. In the following subsections, we discuss the two methods separately.

\subsection{Clothing-Consistent Person ReID}
When the visual appearance of pedestrian clothing does not change, clothing-consistent person ReID methods \cite{Gao2021DCR, xu2020black, he2021transreid} solve short-term pedestrian recognition tasks by focusing on traditional challenges, such as posture, background, and occlusion, and obtain satisfying performance. For example, Ye et al. \cite{Ye2022Deep} divided person ReID systems into closed-world systems and open-world systems. They then conducted a comprehensive review and in-depth analysis of person ReID technology from the standpoints of deep feature representation learning, deep metric learning, and ranking optimization. In addition, the authors designed a powerful AGW (Attention Generalized mean pooling with Weighted triplet loss) baseline and achieved good results. Depending on label usage, clothing-consistent person ReID methods can be divided into unsupervised methods and supervised methods. I) \textbf{Unsupervised methods.} Yu et al. \cite{Yu20Unsupervised} designed a new unsupervised loss function to embed the asymmetry measure into a deep neural network. based on this, they proposed a novel unsupervised deep framework termed deep clustering-based asymmetric metric learning (DECAMEL) to jointly learn feature representations and unsupervised asymmetric metrics. By learning a compact cross-view clustering structure, the view-specific bias was alleviated, and the underlying cross-view discriminative information was mined. Li et al. \cite{Li20Unsupervised} proposed a new unsupervised tracklet association learning (UTAL) framework to solve the person reidentification task. The framework jointly learned intracamera trajectory discrimination and cross-camera trajectory association in a unified architecture to maximize the discovery of trajectory identity matches within camera views and across camera viewpoints. \textbf{ II) Supervised methods.} Many researchers have focused on these methods which can be further divided into image-based person ReID methods and video-based person ReID methods. 1) \textbf{Image-based person ReID methods}. More attention has been paid to image-based person ReID and many image-based person ReID methods have been proposed. For example, Zhou et al. \cite{Zhou22Learning} designed a novel omni-scale network (OSNet) to learn a full-scale feature representation for person ReID that not only captured the features of different spatial scales but also multiscale collaborative combinations. Sun et al. \cite{sun2018beyond} proposed a part-based convolution baseline (PCB) module, where the classic image segmentation method was employed. In the PCB, the feature map was evenly and horizontally segmented to learn the local features, which were combined to ensure that the property of each strip was consistent. This simple and effective approach to unified partitioning has become an important baseline in the field of person ReID. Wang et al. \cite{wang2018learning} proposed a multigranularity network (MGN) that consisted of one global branch and two local branches. The feature learning strategy was applied to obtain the global feature from an entire image and the local features from smaller local areas of the image. Xu et al. \cite{xu2020black} proposed a novel head-shoulder adaptive attention network (HAA), where head-shoulder descriptors were utilized to adaptively solve the person ReID problem. Qian et al. \cite{Qian20Leader} proposed a new two-layer deep network named MuDeep, where deep discriminative feature representations at different scales through a multiscale deep learning layer were learned. Then, a leader-based attention learning layer was utilized to guide the information of multiple scales and to determine the best weight for each scale. Kviatkovsky et al. \cite{Kviatkovsky13Color} suggested that color information as a single clue could obtain good identification properties and by using different parts of the object, the color distribution structure was employed to achieve an invariant signature. Moreover, the intra-distribution structure was utilized as an invariant descriptor, and nonparametric shape descriptors were applied to describe the intra-distribution structure. Li et al. \cite{Li22Pose} proposed a pose-guided representation (PGR) for person ReID, where human pose and partial cues were employed to learn the robustness of pose-invariant features to pose variations and local descriptive features to misalignment errors, respectively. He et al. \cite{he2021transreid} built a new baseline framework named TransReID, where the transformer framework was employed. In this framework, side information such as viewpoint and camera was encoded by learnable embeddings, and rearranging patches were used for local feature learning. Gao et al. \cite{Gao2021DCR} proposed a novel deep spatial pyramid feature collaborative reconstruction module (DCR), where the collaborative reconstruction of different blocks in a query were jointly reconstructed to effectively solve the problems of pedestrian view/pose changes and occlusions. Hou et al. \cite{Hou22Feature} applied feature completion to solve occluded person ReID and designed the region feature completion (RFC) block. Moreover, the spatial and temporal contexts were separately captured to recover the semantic information of the occluded regions in the recovered feature space. 2) \textbf{Video-based person ReID methods}. Some researchers have focused on video-based person ReID methods. For example, Meng et al. \cite{Meng22Deep} constructed spatial and temporal graphs to capture the structure graph information in the original video clip and designed a deep graph metric learning (DGML) method to measure the consistency between the spatial graphs in the video of consecutive frames, where the spatial graph captured the neighborhood relationship about the detected human instances in each frame. Wang et al. \cite{Wang16Person} proposed a discriminative video fragment selection and ranking (DVR) method, which automatically selected the most discriminative video segment from the image sequence of multisegment pedestrians, calculated reliable space-time and appearance features and learned the video ranking function of person ReID.

These methods have achieved good performance in the face of traditional challenges in person ReID. However, the assumption that the visual appearance of clothes is consistent for the same person is needed. When they are directly applied to cloth-changing person ReID tasks (which need to be addressed in real surveillance scenarios), their performance dramatically decreases. Thus, increasing attention is being paid to emerging cloth-changing person ReID tasks. In the following subsection, we introduce this topic.

\subsection{Cloth-Changing Person ReID}

Large-scale datasets play an important role in the optimization of network parameters. Thus, to promote the development of the cloth-changing person ReID task, different cloth-changing person ReID datasets, such as PRCC \cite{yang2021clothing}, LTCC \cite{qian2020long}, Celeb-reID \cite{huang2020beyond}, Celeb-reID-light, \cite{huang2019celebrities}, NKUP \cite{wang2020benchmark}, VC-Clothes \cite{wan2020person}, and COCAS \cite{yu2020cocas}, have been built and released. These datasets provide diverse data thereby facilitating researchers to  assess their proposed modules. Moreover, some cloth-changing person ReID methods have been proposed, whose core idea is to extract pedestrian features that are not related to clothes. For example, Gao et al. \cite{gao2022multigranular} proposed a novel multigranular visual-semantic embedding method (MVSE), where visual semantic information and human attributes are fully explored. Hong et al. \cite{Hong2021FSAM} proposed a fine-grained, shape-appearance mutual learning framework to learn fine-grained, discriminative body shape knowledge. Gu et al. \cite{Gu2022Clothes} proposed clothes-based adversarial loss (CAL) to mine clothing irrelevant features from the original RGB images by penalizing the predictive power of the ReID model. Zhang et al. \cite{Zhang23Specialized} proposed a two-step retrieval verification strategy, where the metric learning results were utilized to filter candidate images, and a local clues-oriented verification network (LCVN) was used to determine the special features of similar images. Moreover, a ranking strategy was introduced to achieve a balance between retrieval results and verification results. Huang et al. \cite{huang2020beyond} used vector neurons instead of scalar neurons to design the ReIDCaps network, where the direction of vector neurons indicated the diversity of clothing information and the length of vector neurons denoted the pedestrian identity information. Shu et al. \cite{Shu2021Semantic} proposed a semantic-guided pixel sampling approach, where the extracted features were not explicitly defined to learn cues unrelated to shirts and pants. Xu et al. \cite{xu2021AFD-Net} proposed an adversarial feature disentanglement network (AFD-Net), where the intraclass feature variation was reduced by intraclass reconstruction and new adversarial dress images were generated by exchanging and recombining cross-identity features. Chen et al. \cite{chen20213DSL} proposed an end-to-end framework for 3D shape learning (3DSL), where 3D human reconstruction was combined from a single image to extract texture-insensitive, 3D features. In this way, 3DSL forces the features to be more robust to texture-confused pedestrians via a regularization method for 3D reconstruction. To solve the problem of low sparsity and diversity in the cloth-changing person ReID task, Jia et al. \cite{Jia22Complementary} designed complementary data augmentation strategies to enhance the feature learning process, including positive and negative data augmentation. Huang et al. \cite{huangRSCANet} explicitly constructed a clothing status awareness learning process and proposed regularization via a clothing status awareness network (RCSANet), which regularized the pedestrian description by embedding clothing status awareness and improving the discriminability of pedestrian features. Chen et al. \cite{chen2022deep} discovered that the human body shape had relatively stronger invariance with moderate clothing changes. Accordingly, they proposed a multiscale appearance and contour deep infomax (MAC-DIM) method. Moreover, the mutual collaboration of color RGB images and contour images was exploited to learn more effective shape-aware and clothing-invariant representations. Zhang et al. \cite{zhang2021unsupervised} designed an unsupervised person reidentification model for changing clothes, named Syn-Person-Cluster ReID, to solve the problem of a lack of data. In the Syn-Person-Cluster ReID, the authors synthesized pedestrian changing images through a changing enhancement model while constraining images of the same person with different clothes to have the same pseudolabel. Jin et al. \cite{jin2022cloth} proposed a two-stream, GI-ReID architecture that consisted of an image ReID stream and auxiliary gait recognition stream. The GI-ReID drove the model to learn a cloth-independent representation by using the unique cloth-independent gait information of pedestrians as a regulation variable. Gu et al. \cite{Gu2022Clothes} proposed a cloth-based adversarial loss (CAL) method that penalized the model's ability to predict clothes by designing a reasonable loss function and mined clothing irrelevant features from the original RGB images. The extracted feature was more robust to changes in clothing. Bansal et al. \cite{bansal2022cloth} proposed a vit-vibe hybrid model to solve the problem of the cloth-changing person ReID task. The model paired unique, soft biometer-based discriminative information with ViT feature representations to learn a robust and unique feature representation invariant to clothing changes. Lee et al. \cite{lee2022attribute} proposed an attribute debiased vision transformer (AD-ViT) method to learn identity-specific features for the cloth-changing person ReID task. Since human appearance exhibits large variations with different clothes, existing methods tend to introduce additional modal cues to assist model learning. For example, SPT+ASE \cite{yang2021clothing} and MAC-DIM \cite{chen2022deep} introduced human body contour sketches to assist learning. On the other hand,  GI-ReID \cite{jin2022cloth} and ViT-VIBE Hybrid \cite{bansal2022cloth} used gait prediction results to assist model learning while 3DSL \cite{chen20213DSL} learns 3D features by introducing 3D human reconstruction regularization. However, the proposed 
IGCL 
uses the idea of collaborative learning to jointly optimize pedestrian features across multiple aspects (i.e., \emph{clothing attention decline}, ii. \emph{strengthen the diversity of posture information}, iii. \emph{highlight strong identity-related local information}), and pay more attention to mining identity-related information from the RGB image itself. Thus, in this work, we design a novel collaborative learning framework that adequately takes advantage of human semantic information and pedestrian identity information and then extracts a generalized and robust feature to represent a person wearing different clothes.

\begin{figure*}[t]
\begin{center}
\includegraphics[width=7in,height = 4.0in]{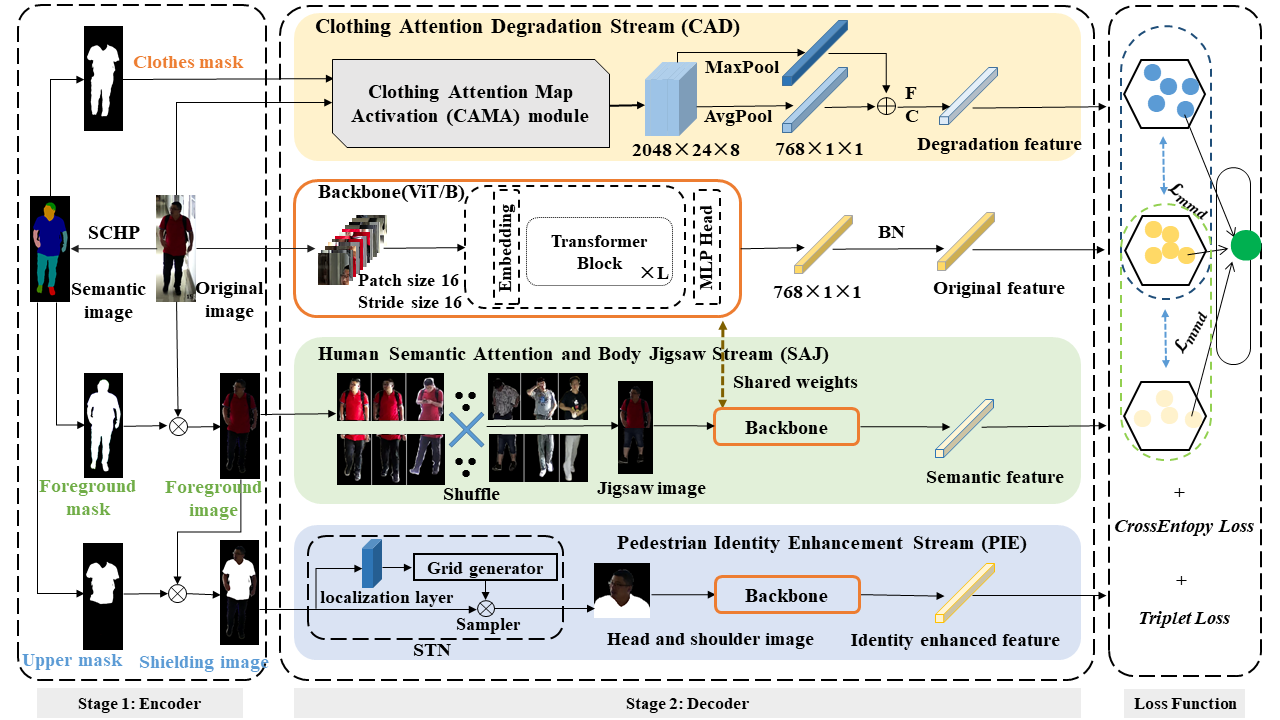}
\vspace{-0.5em}
\caption{Pipeline of the IGCL approach. It consists of the backbone, CAD stream, SAJ stream, and PIE stream, which are jointly optimized in an end-to-end network architecture. We note that in  IGCL, Vision Transformer serves as the backbone. `BN' indicates batch normalization, `STN' denotes the spatial transformer networks, and `SCHP' is a semantic analysis module to obtain human semantic information. `MaxPool' and `AvgPool' indicate maximum pooling and average pooling, respectively. } \label{figure_framework}
\end{center}
\vspace{-1.5em}

\end{figure*}

\section{Methodology}

As illustrated in Figure \ref{figure_framework}, the IGCL method consists of two stages---an encoder and a decoder The loss function is applied to optimize the network parameters of the encoder and decoder. To obtain different and rich representations of the original image, the clothes mask, foreground image, and shielding image are obtained by the encoder. Then these images are fed to the decoder. The decoder consists of the backbone, CAD stream, SAJ stream, and PIE stream, which are collaboratively learned in an end-to-end unified framework. Moreover, the human semantic and identity information is effectively utilized, and the identity is kept unchanged in each stream. Since these streams are complementary, they promote each other, and the extracted feature is more discriminative and robust and the clothing is made irrelevant. We note that the clothing degradation feature (the output of the CAD stream), original feature (the output of the backbone), semantic feature (the output of the SAJ stream), and identity-enhanced feature (the output of the PIE stream) are the outputs of the decoder. In the training phase, all these features are fed to the loss function. However, in the testing phase, only the original feature is utilized to calculate the similarity between the query sample and the gallery sample. In the following subsections, we introduce the encoder, decoder, and loss function.

\begin{figure}[t]
\begin{center}
\includegraphics[width=3.5in,height = 4.0in]{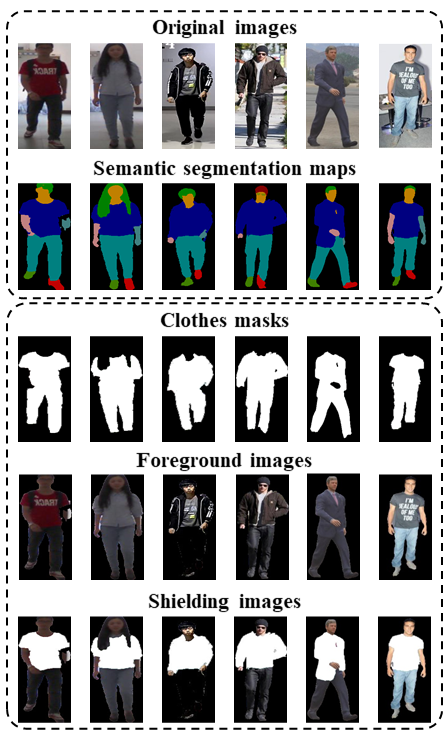}
\caption{Results of the encoder process. From top to bottom are the original images, semantic segmentation maps, clothes masks, foreground images, and shielding images}. \label{encoder process_figure}
\end{center}
\vspace{-1.5em}
\end{figure}

\subsection{Encoder}

Since the same person has different clothes in the cloth-changing person ReID task, the difference in visual appearance is very large, and it is a very challenging job to clearly describe them. To accommodate clothing variations within limited data, more enriched feature representations for each identity are needed. Accordingly, the semantic segmentation model is used in preprocessing to segment body parts, inspired by \cite{gao2022multigranular}\cite{jin2022cloth}. The encoder takes the original image $X \in R^{384 \times 128 \times 3}$ and passes it through the pretrained human parsing SCHP module \cite{Li2022SCHP} to obtain the human semantic image. The image includes 18 semantic parsing parts: 'Background', 'Hat', 'Hair', 'Sunglasses', 'Upper-clothes', 'Skirt', 'Pants', 'Dress', 'Belt', 'Left-shoe', 'Right-shoe', 'Face', 'Left-leg', 'Right-leg', 'Left-arm', 'Right-arm', 'Bag', and 'Scarf'. As a result, based on the semantic image, the clothes mask $M_c = mc_{i, j} \in R^{384 \times 128 \times 1}$, foreground mask $M_f = mf_{i, j} \in R^{384 \times 128 \times 1}$, and upper mask $M_u = mu_{i, j} \in R^{384 \times 128 \times 1}$ can be acquired. In the clothes mask $M_c$, only the clothing area is set to $1$ while other regions of the image are set to $0$. Then, the clothes mask is fed to the CAD stream to reduce the interference caused by the clothing information and to weaken the clothes region information. In the foreground mask $M_f $, all human semantic areas are set to $1$, and other regions are set to $0$. Then, this mask is combined with the original image to obtain the foreground image, where the corresponding pixels of the foreground mask and the original image are multiplied. Then it is input to the SAJ stream. To obtain robust identity features without clothing texture interference, in the upper mask $M_u$, only the upper area of the clothes mask is kept. Then it is combined with the foreground image to obtain the shielding image, where only the upper area of the foreground image is preserved, and its clothes area is set to $1$. The shielding image is fed to the PIE stream. In this way, we obtain four different images for the original image that enrich the feature representations for each identity and are suitable for clothing variations. In Figure \ref{encoder process_figure}, the original image, foreground image, clothes mask, and shielding image are given. From the images, we observe that these images or masks are complementary.

\subsection{Decoder}

The decoder consists of the backbone, CAD stream, SAJ stream, and PIE stream, where these four streams are collaboratively learned in an end-to-end unified framework. Since the vision transformer has obtained good performance on different vision tasks \cite{dosovitskiy2021image}, it is employed as the backbone in our experiments. Its input is the original image $X \in R^{384 \times 128 \times 3}$, and it is divided into 16 fixed size patches $\{ X_i^p \mid p = 1,2, ..., N \}$. In the ViT, $L = 12$ transformer layers are employed to learn the visual characteristics, and then its outputs are further fed to the MLP layer and batch normalization. Thus, the original feature $x_{ori} \in R^{768}$ is extracted to describe a pedestrian. This feature will be further fed to the loss function. we note that ViT is pretrained on the ImageNet dataset, and then the parameters of ViT are further updated by the corresponding cloth-changing person ReID dataset. In the following subsection, we introduce the other three streams: the CAD stream, the SAJ stream, and the PIE stream.

\begin{figure}[t]
\begin{center}
\includegraphics[width=3.5in,height=1.5in]{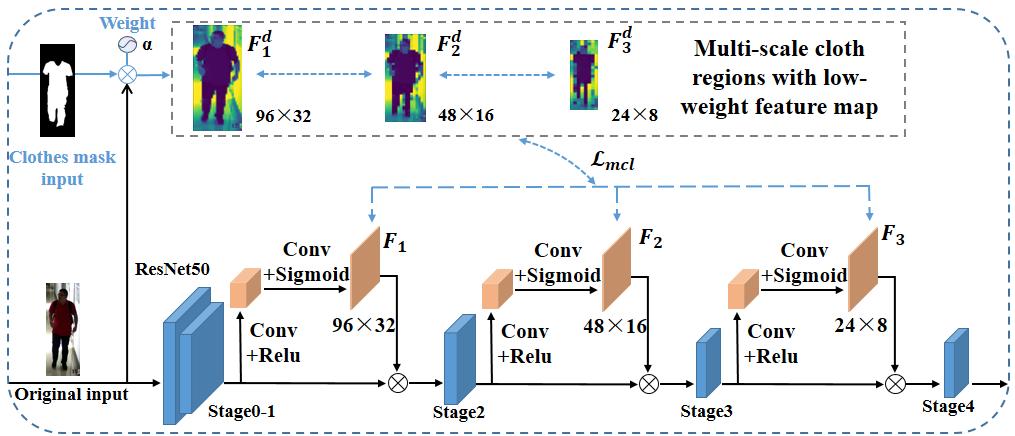}
\vspace{-0.5em}
\caption{Clothing attention map activation (CAMA) module} \label{attention_map_figure}
\end{center}
\vspace{-1.5em}
\end{figure}

\textbf{1) Clothing Attention Degradation Stream (CAD)}. Current works mainly focus on body shape or contour sketches while the clothes region information is not fully explored. We suggest that the clothes region can also provide helpful information. Thus, in this work, a CAD stream is designed to weaken the role of the clothes region and to highlight other unchangeable regions, such as the head, arms, and feet. In this way, the negative effect of the clothes region is reduced to some extent.
The CAD stream consists of a clothing attention map activation (CAMA) module and pooling operations. Through the CAMA module, the CAD stream is gradually guided to reduce the attention to the clothing area and to strengthen the nonclothing clues by using channel attention. Since the convolutional neural network (CNN) and vision transformer network have respective advantages and complementary properties, in the CAMA, ResNet50, which consists of five stages, is selected to extract the feature representations, and ViT is utilized as the backbone of IGCL. In the CAMA, the original image and clothes mask are simultaneously input into the CAD stream. The structure of the CAMA module is given in Figure \ref{attention_map_figure}.

To reduce the interference caused by the clothing information, a low weight value $\alpha$ is given to each pixel in the clothes mask $M_c = mc_{i, j} \in R^{384 \times 128 \times 1}$ and then combined with the original image $X \in R^{384 \times 128 \times 3}$. During fusion, if the pixel does not belong to the clothing region, the pixel value of the original image is kept; if it does belong to the clothing region, the pixel value of the original image is replaced by the corresponding pixel value of the clothes mask. After reducing the weight of all pixels in the clothing area, the multiscale scheme is further employed for the feature maps. Thus, three mid-level feature maps with different scales, $F_1^d$, $F_2^d$, and $F_3^d$, whose dimensions are $96 \times 32$, $48 \times 16$, and $24 \times 8$, respectively, are obtained. In this way, a multiscale clothes mask is obtained where the importance of the clothing area is weakened, and the networks mainly focus on the clothing-irrelevant area. The outline of the human body is reflected by the clothes area to some extent, provides helpful information for feature representation, and prevents the loss of important semantic information due to the rigid coverage of the clothing area.

In addition, the original image $X$ is passed through stages 0-1 of the ResNet50 network to obtain the intermediate feature map $\Phi_1 \in R^{96 \times 32 \times 256}$. Then the intermediate feature map $\Phi_1$ is fed to two $1 \times 1$ convolution layers to obtain the first spatial attention map $F_1 \in R^{96 \times 32}$. Moreover, the intermediate attention map $\Phi_1$ is combined with the first spatial attention map $F_1$. In this way, a new intermediate attention map is obtained and fed to stage 2 of ResNet50. We repeat the above operations three times, and the final weakened attention map $F \in R^{24 \times 8 \times 2048}$ is obtained. The map is defined by

\begin{small}
\begin{equation}
\begin{aligned}
F_i=\sigma \{CV_2*\delta \{ CV_1 * \Phi_i  + b_1 \} +b_2\},
\end{aligned}
\end{equation}
\end{small}
\begin{small}
\begin{equation}
\begin{aligned}
\Phi_i = F_i \otimes \Phi_{i-1} ,
\end{aligned}
\end{equation}
\end{small}
where $F_i \in R^{H_i \times W_i \times 1}$ represents the spatial attention map activated at stage $i$, and $\Phi_i \in R^{H_i \times W_i \times C_i}$ denotes the intermediate feature map of stage $i$. $H_i$ and $W_i$ denote the height and width, respectively, of the spatial feature map $F_i$ and intermediate feature map $\Phi_i$, respectively, which are the same as the mid-level feature map with different scales $F_i^d$. $C_i$ indicates the number of channels of stage $i$. $CV_1$ and $CV_2$ represent two $1 \times 1$ convolution filters, and '*' indicates the convolution operation. $b_1$ and $b_2$ denote the bias, and $\sigma \{ \cdot \}$ and $\delta \{ \cdot \}$ indicate the sigma activation function and ReLU function, respectively. $\Phi_{i-1}$ and $\Phi_i$ indicate the intermediate attention maps of stages $i-1$ and $i$, respectively, and $\otimes$ denotes the Hadamard matrix product. In this way, the final weakened attention map $F \in R^{24 \times 8 \times 2048}$, which is considered to be the output of the CAMA model, is obtained. The attention map $F$ is further enhanced by two pooling functions, and the degradation feature is extracted.

We note that to further make the extracted feature more discriminative and clothing-irrelevant, mid-level collaborative learning schemes (feature distillation learning) between the degradation feature maps with different scales and the spatial attention maps, the output of the CAD stream, and the output of the backbone are utilized. We hope that the difference between the degradation feature and the original feature and that between the degradation feature maps and the spatial attention maps is as small as possible. The details of the feature degradation will be given in the loss function. In addition, ResNet50 is pretrained on the ImageNet dataset, and then the parameters of ResNet50 are further jointly optimized with other networks. Moreover, in the CAD stream, the identity of the original image and the clothes mask is the same; thus, the clothing relevance of the extraction feature is enhanced.

\textbf{2) Human Semantic Attention and Body Jigsaw Stream (SAJ)}. Although human semantic information is very important for robust feature representation in the cloth-changing person ReID task, it is not fully explored in many existing methods. To solve this issue, we propose a novel SAJ stream to highlight human semantic information, simulate different poses of the same identity, and excavate the potential correlation between two feature channels. In this way, the negative effect of the background information is reduced to some extent, and more discriminative features are obtained. In the SAJ stream, the foreground image output by the encoder is fed to the network to highlight the human semantic information. Then the human body jigsaw scheme is applied to the foreground images in the batch. (There are 8 human identities in each batch with each identity having 4 images; thus, in total, there are 32 images in each batch). Specifically, for the top two adjacent images of the same identity in the batch, the upper part of one image and the lower part of another image are exchanged, where both images belong to the same identity with different clothes and different poses (the half body is employed because when the whole body is used, the original image and exchanged image are very different. This makes it very difficult for the feature extractor to describe these differences and obtain discriminative and robust features). Then, the image obtained by exchanging is fed to the ViT, whose network architecture and parameters are the same as that of the backbone, to obtain the semantic feature. To effectively take advantage of the human semantic information, high-level collaborative learning is applied to the original feature and semantic feature (or mutual learning is applied to high-level semantic features), and we hope that the difference between these two features is as small as possible. In this way, human semantic information is not only effectively explored but also partly addresses the problem of pose variations in different views. Moreover, since the human foreground image is employed in the SAJ stream, the extracted features are selectively emphasized to avoid large domain differences after the jigsaw, and they can pay more attention to human semantic information and reduce the negative impact of background information to some extent.

\textbf{3) Pedestrian identity enhancement stream (PIE)}. The current works do not focus on pedestrian identity. As a result, the original image feature is often employed to recognize pedestrian identity, but no special feature is designed for recognizing it. Moreover, we find that both human body features and local features of the head, neck, and shoulder contain identity information, but the latter is more stable than other regions of the human body in the context of the cloth-changing person ReID task. The advantage of the head-neck and shoulder region is that it is a relatively broad region, and the requirement of accurate positioning is not as high as that of the segmentation task. The region can tolerate a slight positioning offset and it will not have a great impact on the performance. Thus, to address this issue, a novel PIE stream is proposed to enhance identity importance, which is employed to identify a pedestrian using head and shoulder information. The structure of the PIE stream is shown in Figure \ref{figure_framework}. In this stream, we adopt the spatial transformer network (STN) \cite{jaderberg2015spatial}, which contains a lightweight localization layer to perform affine transformations on the feature maps to obtain the head and shoulder image with the clothes area set to 1 and the background area set to 0. STN is an ideal choice for the task of extracting local information since it is pretrained to accurately locate the target and perform the affine transformation. Additionally, it is a plug-and-play lightweight module that can be easily integrated into a variety of networks to learn complex spatial transformations. In this way, the head and shoulder image is unrelated to the clothes when the pedestrian has different clothes. Next, the head and shoulder image is fed to the ViT network to obtain the identity-enhanced feature (its network architecture is that of the backbone of IGCL, but the network parameters are not the same and instead are jointly learned with other streams to avoid the overfitting problem). Moreover, the classification loss and triplet loss are utilized for the identity-enhanced feature. Thus, we can better distinguish pedestrians regardless of what clothes they wear and enhance pedestrian identity importance.

\subsection{Loss Function}

To guide the optimization of the network parameters of IGCL, the discriminative loss function is designed. Since the person ReID task is often regarded as a person classification problem, the classification loss is calculated. To further improve the feature discrimination, the triplet loss is added to narrow the intraclass distance and increase the interclass distance. Moreover, to reduce the interference caused by clothing
information, a mid-level collaborative learning scheme between the clothing feature maps and the spatial attention maps is performed. To make the extraction feature pay more attention to the pedestrian identity and to make the information unrelated to clothing, high-level collaborative learning schemes are utilized where the distance metric is employed to measure the probability distribution of different features learned by different streams. Thus, the total loss function is defined by

\begin{small}
\begin{equation}
\begin{aligned}
\mathcal{L}_{total} = \lambda_1 \mathcal{L}_{cls} + \lambda_2 \mathcal{L}_{tri} + \lambda_3 \mathcal{L}_{mcl} + \lambda_4 \mathcal{L}_{hcl},
\end{aligned}
\end{equation}
\end{small}
where $\mathcal{L}_{total}$ is the total loss function of IGCL and $\mathcal{L}_{cls}$, $\mathcal{L}_{tri}$, $\mathcal{L}_{mcl}$, and $\mathcal{L}_{hcl}$ indicate the classification loss, triplet loss, mid-level collaborative learning loss, and high-level collaborative learning loss, respectively. $\lambda_1$, $\lambda_2$, $\lambda_3$, and $\lambda_4$ are tradeoff parameters to balance the contribution of each item; they are empirically set to 1.

For classification loss, common cross-entropy loss is employed. To effectively take advantage of the identity information and better recognize a pedestrian, the original feature $x_{ori} \in R^{768}$, the semantic feature $x_{sem} \in R^{768}$, and the identity enhanced feature $x_{pie} \in R^{768}$ are used to classify the pedestrian. Note that the degradation feature, original feature, semantic feature, and identity-enhanced feature must be passed through a fully connected layer. The Softmax function is then used to calculate the prediction probability, and the cross-entropy loss is computed by comparing the predicted probability distribution with the true probability distribution of the target class, which is one-hot encoded. Classification loss is calculated by:

\begin{small}
\begin{equation}
\begin{aligned}
\mathcal{L}_{cls} = \mathcal{L}_{cls}^{ori} + \mathcal{L}_{cls}^{sem} + \mathcal{L}_{cls}^{pie} + \mathcal{L}_{cls}^{deg},
\\= - \frac{1}{B} \sum_{i=1}^{B} \log p(x_{ori}(i) \mid y_{i})\\
- \frac{1}{B} \sum_{i=1}^{B} \log p(x_{sem}(i) \mid y_{i})\\
- \frac{1}{B} \sum_{i=1}^{B} \log p(x_{pie}(i) \mid y_{i})\\
- \frac{1}{B} \sum_{i=1}^{B} \log p(x_{deg}(i) \mid y_{i}),
\end{aligned}
\end{equation}
\end{small}
where $B$ is the batch size. $p(x_{ori}(i) \mid y_{i})$, $ p(x_{sem}(i) \mid y_{i})$, $ p(x_{pie}(i) \mid y_{i})$, and $p(x_{deg}(i) \mid y_{i})$ are the prediction probabilities of the $i^{th}$ sample belonging to the ground truth $y_i$ for the original feature $x_{ori}$, semantic feature $x_{sem}$, identity enhanced feature $x_{pie}$, and degradation feature $x_{deg}$, respectively.


To further enhance the discrimination of the extraction features, the triple loss is utilized for the original feature $x_{ori} \in R^{768}$ and the semantic feature $x_{sem} \in R^{768}$. Specifically, we randomly select an image as the anchor in the batch and then sample a positive sample image whose label is the same as the anchor and a negative sample image whose label is different from the anchor. Moreover, we hope that the distance between the anchor and the positive sample image is as small as possible but that the distance between the anchor and the negative sample image is as large as possible. The distance is defined as follows:

\begin{small}
\begin{equation}
\begin{aligned}
&\mathcal{L}_{Tri} = \mathcal{L}_{tri}^{ori} + \mathcal{L}_{tri}^{pie}\\ &=\frac{1}{B} \sum_{i=1}^{B}\max \left\{m+d\left(x_{ori}(i), x_{ori}^{pos}(i)\right)-d\left(x_{ori}(i), x_{ori}^{neg}(i)\right), 0\right\}+\\& \frac{1}{B} \sum_{i=1}^{B}\max \left\{m+d\left(x_{sem}(i), x_{sem}^{pos}(i)\right)-d\left(x_{sem}(i), x_{sem}^{neg}(i)\right), 0\right\},
\end{aligned}
\end{equation}
\end{small}
where $m$ is the margin of the triplet loss, which is used to control the difference between the distance of the positive image pair and the distance of the negative image pair ($m$ is set to 0.3 in our experiments). $x_{ori}^{pos}$ and $x_{ori}^{neg}$ indicate the original feature of the positive sample and the original feature of the negative sample, respectively. Similarly, $x_{sem}^{pos}$ and $x_{sem}^{neg}$ indicate the semantic feature of the positive sample and the original feature of the negative sample, respectively. $d(*,*)$ denotes the Euclidean distance.

In the CAD stream, the CAMA module is designed to obtain more discriminative features that are clothing-irrelevant, to reduce the attention to the clothing area, and to strengthen the nonclothing clues by using channel attention. To address this issue, mid-level collaborative learning loss between the clothing feature maps and the spatial attention maps is performed; it is calculated by:

\begin{small}
\begin{equation}
\begin{aligned}
\mathcal{L}_{mcl}= \sum_{k=1}^K( \frac{1}{h_k \cdot w_k} \sum_{i=1}^{h} \sum_{j=1}^{w}({F_k}{(i, j)}-{F_k^d}{(i, j)})^{2}),
\end{aligned}
\end{equation}
\end{small}
where $K$ indicates the number of feature maps at different scales. $F_k$ is the $k^{th}$ spatial attention map, and $F_k^d$ is the $k^{th}$ clothing feature map. $h_k$ and $w_k$ are the height and width, respectively, of the $k^{th}$ feature matrix.


IGCL consists of the backbone, CAD stream, SAJ stream, and PIE stream, and these streams are complementary. Thus, the high-level collaborative learning loss between different streams is employed, where high-level features of these streams are used and the maximum mean discrepancy between different domains in the probability distribution is calculated by the maximum mean discrepancy (MMD) \cite{gretton2012kernel}. The MMD is defined as:
\begin{small}
\begin{equation}
\begin{aligned}
&\mathcal{L}_{hcl}=\mathcal{L}_{mmd}^{sem}+\mathcal{L}_{mmd}^{deg}\\ &=\|\mu(x_{sem})-\mu(x_{ori})\|_{2}^{2}+\|\sigma(x_{sem})-\sigma(x_{ori})\|_{2}^{2}\\ &+\|\mu(x_{deg})-\mu(x_{ori})\|_{2}^{2}+\|\sigma(x_{deg})-\sigma(x_{ori})\|_{2}^{2},
\end{aligned}
\end{equation}
\end{small}
where $x_{sem}$, $x_{ori}$, and $x_{deg}$ are the semantic feature, the original feature, and the degradation feature, respectively. $\mu(\chi)$ and $\sigma(\chi)$ indicate the mean and variance calculation functions, respectively. In this way, we can guide the backbone network by utilizing these high-level semantic features. This approach enables the backbone to focus more on the pedestrian body and extract more discriminative features that are weakly correlated with clothing.

\section{Experiments and Discussion}

To evaluate the performance of the IGCL method, we performed experiments using six public cloth-changing person ReID datasets: LaST \cite{shu2022large}, PRCC \cite{yang2021clothing}, LTCC \cite{qian2020long}, Celeb-reID-light \cite{huang2019celebrities}, NKUP \cite{wang2020benchmark}, and VC-Clothes \cite{wan2020person}. As the cloth-changing person ReID task is a new and challenging research topic, to the best of our knowledge, there have been no comprehensive experiments on any cloth-changing ReID algorithms that use all six cloth-changing person ReID datasets. Therefore, our work is the first to systematically and comprehensively assess algorithmic performance in the context of these six datasets. The remainder of this section is organized as follows: 1) six public cloth-changing person ReID datasets are introduced, 2) the competing methods  used in our experiments are listed, 3) the implementation details are described, and 4) the performance evaluations and comparisons based on these six public datasets are described.

\subsection{Datasets}
Our experiments utilized six datasets, namely,  Celeb-reID-light, PRCC, LTCC, NKUP, VC-Clothes, and LaST. The Celeb-reID-light dataset was sourced from the internet, while the PRCC, LTCC, and NKUP datasets were assembled using images captured by real surveillance cameras. The VC-Clothes dataset was created synthetically from high-definition game footage. The LaST dataset was gathered from movies. Additionally, the PRCC, LTCC, LaST, and VC-Clothes datasets contain both cloth-consistent and cloth-changing data, while the NKUP and Celeb-reID-light datasets contain only cloth-changing data. Note that for privacy reasons, all faces in the NKUP dataset are masked. For more details on these datasets, please refer to Table \ref{dataset}.

\begin{table*}
\fontsize{10}{10}\selectfont
\caption{Statistics of existing long-term, image-based, ReID datasets with clothing changes. Note that 'SC' and 'CC' indicate cloth-consistent data and cloth-changing data, respectively.}
\renewcommand{\arraystretch}{1.2}
\vspace{-1.0em}
\begin{center}
\setlength{\tabcolsep}{2.0mm}
{
\begin{tabular}{c|c|c|cc|c|c|c}
\hline
\multirow{2}{*}{Dataset} & \multirow{2}{*}{Source} & \multirow{2}{*}{Train(ID/Image)} & \multicolumn{2}{c|}{Test(ID/Image)} & \multirow{2}{*}{Cameras} & \multirow{2}{*}{Time Range} & \multirow{2}{*}{Data Style} \\ 
\cline{4-5} &   &  & \multicolumn{1}{c|}{Query}    & Gallery     &     &     &    \\ \hline
PRCC \cite{yang2021clothing}                & Real      & 150/17,896   & \multicolumn{1}{c|}{71/3,543} & 71/3,384  & 3     & -         & SC/CC  \\ 
LTCC \cite{qian2020long}                & Real      & 77/9,576     & \multicolumn{1}{c|}{75/493}   & 75/7,050  & 12    & 2 months  & SC/CC  \\ 
Celeb-reID-light \cite{huang2019celebrities}    & Internet  & 490/9,021    & \multicolumn{1}{c|}{100/887}  & 100/934   & -     & -         & CC     \\ 
NKUP \cite{wang2020benchmark}                 & Real      & 40/5,336     & \multicolumn{1}{c|}{39/332}   & 67/4,070  & 15    & 4 months  & CC     \\ 
VC-Clothes \cite{wan2020person}          & Synthetic & 256/9,449    & \multicolumn{1}{c|}{256/1020} & 256/8,591 & 4     & N/A       & SC/CC  \\ 
LaST \cite{shu2022large}          & Movies & 5,000/71,248    & \multicolumn{1}{c|}{5,805/10,176} & 5,806/125,353 & -     & N/A       & SC/CC  \\ 
\hline
\end{tabular}
}
\end{center}
\vspace{-1.0em}
\label{dataset}%
\end{table*}

\subsection{Competitors}
The task of cloth-changing person ReID is a new and challenging topic that has also aroused the interest of researchers in related fields in the last 2-3 years. In our experiments, the latest and popular references were utilized as our competitors, including ReIDCaps (TCSVT2020) \cite{huang2020beyond}, Pixel Sampling (ISPL 2021) \cite{Shu2021Semantic}, AFD-Net (IJCAI 2021) \cite{xu2021AFD-Net}, 3DSL (CVPR 2021) \cite{chen20213DSL}, FSAM (CVPR 2021) \cite{Hong2021FSAM}, RCSANet (ICCV 2021)\cite{huangRSCANet}, MAC-DIM (TMM 2022)\cite{chen2022deep}, Syn-Person-Cluster (ISPL 2022)\cite{zhang2021unsupervised}, GI-ReID (CVPR 2022) \cite{jin2022cloth}, CAL (CVPR 2022) \cite{Gu2022Clothes}, 3APF (CVPRW2020) \cite{wan2020person}, MVSE (ACM MM 2022) \cite{gao2022multigranular}, ViT-VIBE Hybrid (WACV 2022) \cite{bansal2022cloth}, SPT+ASE (TPAMI 2021) \cite{yang2021clothing}, Re-Rank+LCVN (PR 2023) \cite{Zhang23Specialized}, Pos-Neg (TIP 2022) \cite{Jia22Complementary}, and AD-ViT (AVSS 2022) \cite{lee2022attribute}. Additionally, in the cloth-changing person ReID task, traditional person ReID algorithms, such as ResNet50 (CVPR 2016) \cite{he2016deep}, Vision Transformer (ICLR 2021) \cite{dosovitskiy2021image}, PCB (ECCV 2018) \cite{sun2018beyond}, and MGN (ACM MM 2018) \cite{wang2018learning}, are often employed. In our experiments, we also compared IGCL with them. 
More  information about these competitors can be obtained in the related work section.

\begin{table*}
\fontsize{7}{7}\selectfont
\caption{Performance evaluation and comparison on six public cloth-changing person ReID datasets. Bold values indicate the best performance of different backbones in each column, and underline values represent the overall best performance in each column. The '*' indicates results reproduced using the ViT backbone through the source code.} 
\renewcommand{\arraystretch}{1.2}
\vspace{-1.0em}
\begin{center}
\setlength{\tabcolsep}{2.2mm}{
\begin{tabular}{c|c|cc|cc|cc|cc|cc|cc}
\hline

\multirow{3}{*}{Methods} &\multirow{3}{*}{Backbones} &\multicolumn{12}{c} {Datasets} \\ \cline{3-14} 
& & \multicolumn{2}{c|} {PRCC} & \multicolumn{2}{c|} {LTCC} & \multicolumn{2}{c|} {Celeb-reID-light} & \multicolumn{2}{c|} {NKUP} & \multicolumn{2}{c|} {VC-Clothes} & \multicolumn{2}{c} {LaST}\\ \cline{3-14}
& &mAP &rank-1 &mAP &rank-1 &mAP &rank-1 &mAP &rank-1 &mAP &rank-1 &mAP &rank-1\\ \hline

PCB \cite{sun2018beyond} &ResNet &- &22.9 &8.8 &21.9  &- &- &14.1 &18.7 &- &- &15.2 &50.6 \\
MGN \cite{wang2018learning} &ResNet &- &25.9 &10.1 &24.2  &13.9 &21.5 &\textbf{16.1} &\textbf{20.6} &- &- &17.6  &41.0 \\ 
ResNet-50 \cite{he2016deep} &ResNet &8.1 &19.6 &8.4 &20.7 &6.0 &10.3 &4.8 &9.6 &47.4 &50.1 &- &- \\
Pixel Sampling \cite{Shu2021Semantic} &ResNet &61.2 &\textbf{\uline{65.8}} &16.1 &42.3 &14.2 &24.5 &8.5 &13.9 &63.7 &68.2 &- &- \\
AFD-Net \cite{xu2021AFD-Net} &ResNet &- &42.8 &- &- &11.3 &22.2 &- &- &- &- &- &- \\
3DSL \cite{chen20213DSL} &ResNet &- &51.3 &14.8 &31.2 &- &- &- &- &81.2 &79.9 &- &- \\
FSAM \cite{Hong2021FSAM} &ResNet &- &54.5 &16.2 &38.5 &- &- &- &- &78.9 &78.6 &- &- \\
MAC-DIM \cite{chen2022deep} &ResNet &- &48.8 &13.0 &29.9 &- &- &- &- &80.0 &82.0 &- &- \\
SPT+ASE \cite{yang2021clothing} &ResNet &- &34.4 &- &- &- &- &- &- &- &- &- &- \\
Syn-Person-Cluster \cite{zhang2021unsupervised} &ResNet &39.8 &43.7 &- &- &- &- &- &- &62.5 &67.4 &- &- \\
GI-ReID \cite{jin2022cloth} &ResNet &- &37.6 &14.2 &28.9 &- &- &- &- &57.8 &64.5 &- &- \\
CAL \cite{Gu2022Clothes} &ResNet &55.8 &55.2 &18.0 &40.1 &- &- &- &- &81.7 &81.4 &\textbf{28.8} &\textbf{\uline{73.7}} \\
Re-Rank+LCVN \cite{Zhang23Specialized} &- &62.6 &64.4 &- &- &- &- &- &- &\textbf{\uline{86.6}} &87.9 &- &- \\
Pos-Neg \cite{Jia22Complementary} &ResNet &\textbf{\uline{65.8}} &54.9 &14.4 &36.2 &- &- &- &- &- &- &- &- \\ 

3APF \cite{wan2020person} &ResNet &- &- &- &- &- &- &- &- &82.1 &\textbf{\uline{90.2}} &- &- \\ 

\textbf{IGCL} (ours)  &ResNet 
&57.4 &57.8 
&\textbf{35.0} &\textbf{62.0} 
&\textbf{16.2} &\textbf{23.8} 
&14.7 &18.2 
&83.6 &82.0 
&17.6 &40.9 \\ 
\hline
DenseNet-121 \cite{huang2017densely} &DenseNet &23.7 &18.7 &10.7 &27.2 &5.3 &10.5 &10.7 &15.4 &66.8 &67.5 &- &- \\
ReIDCaps \cite{huang2020beyond}  &DenseNet &- &- &- &- &\textbf{19.0} &\textbf{33.5} &- &- &- &- &- &- \\
RCSANet \cite{huangRSCANet} &DenseNet &31.5 &31.6 &- &- &16.7 &29.5 &- &- &- &- &- &- \\
MVSE \cite{gao2022multigranular} &DenseNet &52.5 &47.4 &\textbf{33.0} &\textbf{70.5} &- &- &\textbf{17.9} &\textbf{23.8} &- &- &- &- \\
\textbf{IGCL} (ours)  &DenseNet 
&\textbf{59.2} &\textbf{63.1} 
&31.4 &60.5 
&17.4 &25.4 
&17.8 &20.9 
&\textbf{83.5} &\textbf{81.6} 
&\textbf{20.8} &\textbf{47.6} \\
\hline
baseline \cite{dosovitskiy2021image} &ViT  &46.4 &46.3 &28.6 &69.5 &17.1 &30.2 &11.6 &17.3 &70.6 &71.2 &12.0 &47.0 \\ 
ViT-VIBE Hybrid \cite{bansal2022cloth} &ViT &- &47.0 &38.3 &73.6 &- &- &- &- &- &- &- &- \\
AD-ViT \cite{lee2022attribute} &ViT &- &- &34.2 &72.0 &- &- &16.9 &23.6 &- &- &- &- \\
CAL* \cite{Gu2022Clothes}   &ViT  &33.6 &32.7 &9.2 &18.1 &- &- &- &- &52.5 &49.2 &13.2 &51.1 \\
MVSE*  \cite{gao2022multigranular}  &ViT &28.8 &27.3 &16.8 &39.3 &- &- &8.5 &11.2 &- &- &- &- \\
\textbf{IGCL} (ours)  &Vision Transformer
&\textbf{63.0} &\textbf{64.4}
&\textbf{\uline{47.1}} &\textbf{\uline{77.8}} 
&\textbf{\uline{25.6}} &\textbf{\uline{38.7}} 
&\textbf{\uline{28.0}} &\textbf{\uline{28.8}} 
&\textbf{85.4} &\textbf{82.9}
&\textbf{\uline{39.5}} &\textbf{67.7} \\
\hline
\end{tabular}}
\end{center}
\vspace{-1.0em}
\label{Performance}%
\end{table*}

\subsection{Implementation Details}
In our experiments, the vision transformer (ViT) \cite{he2021transreid}, including an additional batch-normalization bottleneck layer, served as the backbone of the proposed IGCL (it is also considered to be the baseline). ViT was pretrained on the ImageNet dataset, and then the training samples of the PRCC, LTCC, Celeb-reID-light, NKUP, and VC-Clothes datasets were separately employed to fine-tune the modules, including ViT and IGCL. Note that the default cloth-changing settings and divisions of these datasets \cite{yang2021clothing}, \cite{qian2020long}, \cite{zhong2017re}, \cite{huang2019celebrities}, \cite{wang2020benchmark}, \cite{wan2020person}) were applied. In the training procedure, the minibatch size was set to $32$. It contained $8$ pedestrian identities with $N=4$ images per identity, and the input person images were resized to $ 384 \times128 $. In the optimization process, the stochastic gradient descent (SGD) optimizer was employed with a momentum of $0.9$ and a weight decay of $5e^{-4}$, and the model was trained for $60$ epochs. The learning rate was initialized as $7.0 \times 10^{-4}$ with cosine learning rate decay. The hyperparameters $\alpha$ were empirically set to 0.1. Note that the CAD stream, SAJ stream, and PIE stream were used only in the training stage to jointly optimize the network parameters of the backbone. On the other hand,  in the test stage, only the backbone was used to extract the feature representation, where the backbone focused on extracting features that are not related to clothes and are more general. Thus, only the original features extracted by the backbone from the original RGB image were used to describe each person. In addition, in the optimization and inference phases, the Euclidean distance normalized by the L2 norm was always utilized to calculate the similarity between any two images.

\textbf{Evaluation Protocols}. In the person ReID community, the cumulative matching characteristic (CMC) curves, rank-1 and mean average precision (mAP) are often utilized as the evaluation metrics \cite{yang2021clothing, Gao2021DCR, jin2022cloth}. Accordingly, we also utilize these metrics in our experiments.

\subsection{
Performance evaluations and comparisons}

In this section, we report the results obtained by evaluating the performance of IGCL on six public changer ReID datasets and comparing it with the abovementioned competitors. For the open-source code algorithms, ImageNet was utilized to pretrain their backbones. Then the training samples of the LaST, LTCC, PRCC, Celeb-reID-light, NKUP, and VC-Clothes datasets were employed for fine-tuning. The test samples of each of the six cloth-changing person ReID datasets were used to assess their performances. If the same datasets were used in the comparison, we directly cite the reported results. If the source code is released but some datasets were unused, we followed the original paper settings, ran the source code on unused datasets, and report the results. For CAL and MVSE, the original backbone was replaced by others such as ResNet, DenseNet, or VisionTransformer. Note that for GI-ReID \cite{jin2022cloth}, we chose the highest results among their multiple baselines. Pixel sampling \cite{Shu2021Semantic} on the PRCC dataset used original paper results, while other datasets were from our code reproduction. CAL* and MVSE* were from our code replication with the ViT backbone. The results are shown in Table \ref{Performance}. From these results, we obtain the following observations:

1) Regardless of which method is selected, IGCL achieves the best performance on the LTCC, Celeb-reID-light, and NKUP datasets, with significant improvement in mAP and rank-1 over existing algorithms. For other datasets, our proposed IGCL still achieves comparable performance. For example, the mAP and rank-1 accuracy of IGCL on the LTCC dataset are 47.1\% and 77.8\%, respectively, while the corresponding mAP and rank-1 accuracy of the baseline reach 28.6\% and 69.5\%. The largest improvements are 18.5\% (mAP) and 8.3\% (rank-1). Similarly, when the NKUP dataset is utilized, the mAP and rank-1 of IGCL are 28\% and 28.8\%, respectively, while the corresponding performance of the baseline is 11.6\% and 17.3\%, and its corresponding improvement reaches 16.4\% (mAP) and 11.5\% (rank-1). In addition, when the LaST dataset is employed, IGCL can obtain comparable performance. Thus, IGCL significantly outperforms the baseline. Joint optimization of the IGCL model by embedding CAD, SAJ, and PIE streams into the Vision Transformer gives the model the ability to extract features with strong discrimination. Moreover, the model effectively weakens the clothing region information and highlights the robust pedestrian identity features. Most importantly, the model strives to mine the visual semantic information obtained from RGB images to learn a pedestrian identification representation that is invariant to clothing changes.

When compared with specifically designed clothing-changing person ReID methods \cite{huang2020beyond}-\cite{lee2022attribute}, ReIDCaps achieves the second-place performance on the Celeb-reID-light dataset, whose mAP and rank-1 are 19.0\% and 33.5\%, respectively. When compared with IGCL, the corresponding performance improvement reaches 6.6\% (mAP) and 5.2\% (rank-1). When the LTCC dataset is applied, the mAP and rank-1 of the ViT-VIBE Hybrid are 38.3\% and 73.6\%, respectively while the corresponding mAP and rank-1 of the IGCL are 47.1\% and 77.8\%, with improvements  8.8\% and 4.2\%, respectively. When using the VC-Clothes dataset, 3APF, and IGCL achieve mAP/Rank-1 of 82.1\%/90.2\% and 83.6\%/82\%, respectively. IGCL outperforms 3APF's method in mAP accuracy but has a lower rank-1. When the PRCC dataset is used, the mAP/Rank-1 s of Pos-Neg and IGCL are 65.8\%/54.9\%, and 63\%/64.4\%, respectively. IGCL has a higher Rank-1 accuracy than Pos-Neg's method and a decrease of 2.8\% compared to its mAP. However, when the LTCC dataset is used, the mAP/Rank-1 s of Pos-Neg and IGCL are 14.4\%/36.2\%, and 47.1\%/77.8\%, respectively. IGCL can obviously outperform Pos-Neg's method. For Rerank+LVCN, different backbones are suitable for different datasets. When ADB-Net is used as the backbone, Rerank+LVCN can obtain the first-place performance on the VC-Clothes dataset, but it can only obtain the third-place performance on the PRCC dataset. Similarly, on the PRCC dataset, Rerank+LVCN with FlipReID can obtain the best performance. Thus, it is very difficult to select a suitable backbone for different datasets. These specially designed methods mainly focus on contour sketching or modeling human body shapes from multimodality information to avoid interference caused by clothing information. However, directly using this information will produce large semantic loss, lose important identity features, and cannot effectively exploit the complex background and human body semantic information.

We also compared IGCL with GI-ReID. In the latter, the gait information of pedestrians is employed and drives the person ReID model to learn the representation independent of clothes. Since the human semantic information is not fully explored and the predicted gait results are not always perfect, the performance is significantly affected. However, in  IGCL, CAD, and SAJ streams are designed to guide the model to pay more attention to the features that are independent of clothes and to focus on the human semantic information, while the additional PIE stream forces the model to extract more favorable identity robust biometric features. The experimental results show that IGCL has good generalization ability and that the trained model reduces the negative effects caused by background and clothing changes to some extent. Overall, IGCL consistently outperforms all these SOTA methods on five public datasets which demonstrates the effectiveness of the IGCL.

2) Among the clothing-consistent person ReID methods, the MGN achieves the best performance regardless of which dataset is selected. Therefore, we compared MGN with IGCL with ViT on different datasets. On the PRCC dataset, the rank-1 s of MGN and IGCL are 25.9\% and 64.4\%, respectively, with the improvement reaching 38.5\%. Similarly, when using the LTCC dataset, the improvement in our method reaches 37.0\% (mAP) and 53.6\% (rank-1).  We can draw the same conclusions from other datasets. We observe that the clothing-consistent person ReID methods mainly learn features from the clothing appearance, but the clothes cover most of the image of a person, and their visual appearance must be similar. However, for the challenging long-term CC-ReID task, the clothing appearance information of the person often exhibits large changes. Hence when these methods are directly employed, they cannot perform well. In the task of changing clothes person recognition, the core idea is to mine the clues that are not related to clothes but instead are sensitive to identity. Accordingly, the proposed IGCL extracts feature independent of clothing, and its performance is significantly better than that of clothing-consistent person ReID methods.

3) ResNet50, DenseNet121, and Vision Transformer models are widely utilized in many deep learning tasks but are also often evaluated on person ReID tasks. Although these models achieve good performance in many related tasks, their performance is not ideal when they are directly applied to the cloth-changing person ReID task. For example, when using the PRCC dataset, the rank-1 accuracies of ResNet50, DenseNet121, Vision Transformer, and IGCL with ViT are 19.6\%, 18.7\%, 46.3\%, and 64.4\%, respectively. The corresponding improvements achieved by IGCL are 44.8\%, 45.7\%, and 18.1\%, respectively. The mAP accuracies of ResNet50, DenseNet121, Vision Transformer, and IGCL with ViT on the VC-Clothes dataset are 47.4\%, 66.8\%, 70.6\%, and 85.4\%, respectively. The corresponding improvements achieved by IGCL with ViT are 38\%, 18.6\%, and 14.8\%, respectively. Although these network models are widely employed in different tasks, they cannot address cloth-changing features. In IGCL, the model pays more attention to the features unrelated to clothes thereby reducing the impact of cloth-changing data to some extent. In addition, Vision Transformer achieves much better performance than ResNet50 and DenseNet121 regardless of which dataset is utilized. Therefore, Vision Transformer served as the backbone of the IGCL in our experiments. From Table \ref{Performance}, we can observe that when different backbones are embedded into the IGCL, IGCL with ViT can obtain the best performance regardless of which dataset is utilized.

4) We also compared IGCL with existing works by replacing different backbones. When using ResNet50 or DenseNet121 as the backbone, IGCL's performance was degraded to a certain extent. However, IGCL with VisionTransformer consistently achieved the best performance regardless of the dataset used. We also replicated CAL and MVSE using VisionTransformer as the backbone, resulting in CAL* and MVSE*. Interestingly, when CAL and MVSE used VisionTransformer, their performance showed a cliff-like decline. The reasons are that the difference in performance can be attributed to the varying compatibility levels between the network architectures and the accompanying algorithms. MVSE is specifically optimized to handle local and multi-scale information, while CAL relies on correlating local context to encourage the model to penalize  its local clothing prediction ability. In this regard, the convolution and pooling operations of the CNN backbone align well with these requirements of MVSE and CAL. On the other hand, IGCL primarily focuses on capturing global information and placing emphasis on acquiring global context and long-distance dependencies. In such cases, the global self-attention mechanism of the ViT backbone may offer a more advantageous approach to effectively integrate these types of information, but it cannot well match with the MVSE and CAL. Thus, the IGCL with VisionTransformer can obtain the best performance. 

\section{Ablation Study}
An ablation study was performed using IGCL to analyze the contribution of each component. In our experiments, three representative datasets, including the internet Celeb-reID-light dataset, real PRCC dataset, and synthetic VC-Clothes dataset, were selected to verify the effectiveness of IGCL. In this investigation, five aspects were considered: 1) the effectiveness of the CAD module, 2) the advantages of the SAJ module, 3) the benefits of the PIE module, 4) convergence analysis, 5) complexity analysis, and 6) qualitative visualization. In the following subsection, we discuss these six aspects.

\subsection{Effectiveness of the CAD Module}

In many existing person ReID methods, human shape or gait features are usually employed to resist the interference caused by clothing information. In this section, we assess the effectiveness of the CAD module on three cloth-changing person ReID datasets, where the importance of eliminating clothing-related information is considered. Since ViT is the backbone of IGCL, in our experiments, ViT was applied as the baseline, Softmax was employed as the classification function, and then the CAD stream was embedded in the baseline. Since different scale attention feature maps with restricted clothing areas in the CAD model were employed for distillation learning, we analyzed the effectiveness of the CAD module when single-scale and multiscale attention feature maps were used. Moreover, we verified the rationality of the selection of clothing elimination weight $\alpha$. The results are given in Table \ref{Multi-Scale-CAD} and Table \ref{Weight-CAD}. We make the following observations.

\begin{table}
\caption{Effectiveness of the CAD stream when different scales attention feature maps with restricted clothing area in the CAD stream are employed. $F_1^d$, $F_2^d$, and $F_3^d$ indicate that the attention feature maps unrelated to clothing regions are $96\times32$, $48\times16$, and $24\times8$, respectively, and `+' denotes that these attention feature maps are embedded in the baseline for distillation learning.}
\renewcommand{\arraystretch}{1.4}
\vspace{-1.0em}
\maketitle
\begin{center}
\setlength{\tabcolsep}{1.8mm}{
\begin{tabular}{c|cc|cc|cc}
\hline
\multirow{3}{*}{Methods} &\multicolumn{6}{c} {Datasets} \\
\cline{2-7}
& \multicolumn{2}{c|} {PRCC} &\multicolumn{2}{c|} {Celeb-reID-light} &\multicolumn{2}{c} {VC-Clothes} \\
\cline{2-7}
&mAP &rank-1 &mAP &rank-1 &mAP &rank-1\\
\hline
Baseline                     &46.4 & 46.3 & 17.1 & 30.2 & 70.6 & 71.2 \\
+$F_1^d$                        &51.1 & 53.3 & 21.3 & 37.4 & 71.0 & 71.6 \\
+$F_1^d$+$F_2^d$              &50.4 & 52.9 & \textbf{22.0} & 36.1 & 70.6 & 71.2 \\
\boldmath{+$F_1^d$+$F_2^d$+$F_3^d$}    &\textbf{51.8} & \textbf{54.2} & 21.3 & \textbf{37.8} & \textbf{71.8} & \textbf{72.9} \\
\hline
\end{tabular}}
\end{center}
\vspace{-1.0em}
\label{Multi-Scale-CAD}%
\end{table}

\begin{table}
\caption{Effectiveness of the CAD Stream when different clothing elimination weights $\alpha$ are employed. Note that three scale attention feature maps with restricted clothing areas in the CAD model are employed. }
\renewcommand{\arraystretch}{1.4}
\vspace{-1.0em}
\maketitle
\begin{center}
\setlength{\tabcolsep}{2.2mm}{
\begin{tabular}{c|cc|cc|cc}
\hline
\multirow{3}{*}{Methods} &\multicolumn{6}{c} {Datasets} \\
\cline{2-7}
& \multicolumn{2}{c|} {PRCC} &\multicolumn{2}{c|} {Celeb-reID-light} &\multicolumn{2}{c} {VC-Clothes} \\
\cline{2-7}
&mAP &rank-1 &mAP &rank-1 &mAP &rank-1\\
\hline
0.01             &49.6 & 52.0 & 21.8 & 36.6 & 70.6 & 71.4 \\
\textbf{0.1}     &\textbf{51.8} & \textbf{54.2} & 21.3 & \textbf{37.8} & \textbf{71.8} & \textbf{72.9} \\
0.3              &48.6 & 51.8 & 21.3 & 36.6 & 71.4 & 71.0 \\
0.5              &49.0 & 51.5 & 18.0 & 34.6 & 71.2 & 72.0 \\
0.7              &48.8 & 52.1 & \textbf{21.9} & 37.4 & 71.0 & 71.6 \\

\hline
\end{tabular}}
\end{center}
\vspace{-1.5em}
\label{Weight-CAD}%
\end{table}

1) In our experiments, different scale attention feature maps with restricted clothing areas were added to the baseline step by step. Moreover, the weight $\alpha$ of clothing restriction was set to 0.1, and the results are shown in Table \ref{Multi-Scale-CAD}. Adding feature map $F_1^d$ to the baseline improves performance to varying degrees, as observed in `+$F_1^d$'. For example, when using the PRCC dataset, the mAP/rank-1 of `+$F_1^d$' and the baseline are 51.1\%/53.3\%, and 46.4\%/46.3\%, respectively, whose improvement reaches 4.7\%/7.0\%. Similarly, when the Celeb-reID-light dataset is selected, the mAP/rank-1 of `+$F_1^d$' and the baseline are 21.3\%/37.4\%, and 17.1\%/30.2\%, respectively, whose improvement reaches 4.2\%/7.2\%.  Moreover, when multiscale attention feature maps with restricted clothing areas are employed, their performance is further improved, especially for three-scale attention feature maps. For example, the mAP/rank-1 s of the baseline and `+$F_1^d$+$F_2^d$+$F_3^d$' are 46.4\%/46.3\%, and 51.8\%/54.2\%, respectively, whose improvement reaches 5.4\%/7.9\%. These experimental results prove that the CAD stream is effective, and the regularization constraint on the feature maps of different scales effectively and reasonably reduces the sensitivity to clothing texture. Regarding the VC-Clothes dataset, its improvement is insignificant due to its virtual nature, clear data samples, and easily identifiable identity features. Consequently, the auxiliary enhancement strategy yields a relatively small gain.

2) We also investigated the effectiveness of the weights that restrict the clothing region in the feature map. In our experiments, three scales of attention feature maps with restricted clothing areas in the CAD model were employed. We set $\alpha$ to $\left\{0.01,0.1,0.3,0.5,0.7\right\}$, and the results are given in Table \ref{Weight-CAD}. We observe that when a different $\alpha$ is applied, its performance varies. When $\alpha$ is set to 0.1, its rank-1 is almost the best regardless of the dataset. When the value of $\alpha$ is smaller or larger, its performance decreases. Since most of the datasets consist of low-resolution images and the clothing area labeled by the human body parsing model may not be accurate. Hence the clothing area cannot be accurately eliminated without affecting the other areas.  As a result, it is easy to lose important visual semantic information if the clothing area is too restricted, and it is advisable to set a small value for $\alpha$. Thus, in our experiments, the weight $\alpha$ is empirically set to 0.1.  It is important to note that setting $\alpha$, to 0.1 is not a definitive factor, as it may not necessarily be the best choice. When $\alpha$ is assigned another suitable value, the CAD technique still demonstrates varying degrees of improvement.

\subsection{Advantages of the SAJ Module}
We next verified the advantages of the SAJ module on three public cloth-changing person ReID datasets. The experimental results are shown in Table \ref{RGJ-Table}. Note that the ViT backbone is treated as the baseline when the original image is fed to the Vision Transformer. In addition, when the foreground semantic image (different parts of the human body) is fed to the baseline, these two streams are jointly optimized, and their results are named `+Semantic'. Moreover, when the human body jigsaw scheme is used in the SAJ module, we name it `+Semantic+Jigsaw'. When only the foreground semantic image of the SAJ stream is added to the baseline, its performance can still exhibit a large improvement. For example, when using the Celeb-reID-light dataset, the mAP and rank-1 accuracy of '+Semantic' are 22.2\% and 38.6\%, respectively, and the mAP and rank-1 accuracy of the baseline are 17.1\% and 30.2\%, respectively, whose improvement reaches 5.1\% (mAP) and 8.4\% (Rank-1). On the PRCC and VC-Clothes datasets, the rank-1 accuracies of '+Semantic' are 63.4\% and 76.5\%, respectively, which are 17.1\% and 5.3\% higher, respectively, than the baseline. Since the foreground semantic image and distillation learning are used for the baseline and the SAJ stream, it can highlight the human semantic information, and the negative effect of the background information can be reduced to some extent. Moreover, more discriminative features are obtained. In addition, when the human body jigsaw scheme is utilized, its performance is further improved. Unfortunately, there are several existing challenges that hinder the effectiveness of enhancement strategies and restrict their full utilization when dealing with biased foreground images. These challenges include small-sized and low-quality images, as well as inaccurate semantic segmentation. However, it is important to highlight that this approach can moderately simulate clothing variations and different viewpoints/poses of the same identity. This capability enhances the model's adaptability to human pose changes in most cases and improves the overall algorithm performance.

\begin{table}
\caption{Effectiveness of the human semantic information and the body jigsaw in the SAJ stream}
\renewcommand{\arraystretch}{1.4}
\vspace{-1.0em}
\begin{center}
\setlength{\tabcolsep}{1.2mm}{
\begin{tabular}{c|cc|cc|cc}
\hline
\multirow{3}{*}{Methods} &\multicolumn{6}{c} {Datasets} \\
\cline{2-7}
& \multicolumn{2}{c|} {PRCC} &\multicolumn{2}{c|} {Celeb-reID-light} &\multicolumn{2}{c} {VC-Clothes} \\
\cline{2-7}
&mAP &rank-1 &mAP &rank-1 &mAP &rank-1\\
\hline
Baseline   & 46.4 & 46.3 & 17.1 & 30.2 & 70.6 & 71.2 \\
+Semantic & \textbf{59.6} & 63.4 & 22.2 & 38.6 & 76.3 & 76.5 \\
+Semantic+Jigsaw & 59.5 & \textbf{63.5} & \textbf{22.8} & \textbf{39.5} & \textbf{76.6} & \textbf{77.3} \\ 
\hline
\end{tabular}}
\end{center}
\vspace{-1.5em}
\label{RGJ-Table}
\end{table}

\subsection{Benefits of the PIE Module}
We then evaluate the effectiveness of the PIE stream on three publicly available cloth-changing person ReID datasets. As mentioned in Part A of Section V, the ViT is used as the baseline where the Softmax is employed as the classification function. Then the PIE stream is  embedded in the baseline. However, in the PIE stream, different images, including the original image, foreground image, and shielding image, are employed, and then the partial head, neck, and shoulder of these images are obtained to  feed the ViT of the PIE stream. The results are shown in Table \ref{PIE-Stream-Table}, where `+PIE [Orig]', `+PIE [Orig w/ Shielding]', and `+PIE [FG w/ Shielding]' indicate that the PIE stream is combined with the baseline, but the original image, original image with shielding, and foreground image with shielding, respectively, are utilized. We determine that:

1) When the PIE stream is embedded in the baseline, the performance can be improved regardless of which kind of image is selected. For example, when using Celeb-reID-light, the mAP/rank-1 s of the baseline and `+PIE [Orig]' are 17.1\%/30.2\%, and 22.0\%/36.9\%, respectively, and its improvement is 4.9\% (mAP) and 6.7\% (rank-1). Similarly, when the VC-Clothes dataset is employed, the improvement in `+PIE [Orig]' reaches 2.7\% (mAP) and 1.7\% (rank-1), respectively. Thus, these  results show that the PIE stream is very effective and very useful for enhancing identity information. Moreover, when the shielding strategy where the upper clothes area is covered is  applied for the original image or foreground image, its performance is further improved. For example, when compared with `+PIE [Orig]', the mAP and rank-1 of `+PIE [Orig w/ Shielding]' on the VC-Clothes dataset are further improved by 2.1\% and 2.2\%, respectively. When the PRCC dataset is employed, the mAP accuracy and rank-1 accuracy of `+PIE [FG w/ Shielding]' and `+PIE [Orig w/ Shielding]' are 49.3\%/52.6\%, and 47.0\%/47.3\%, respectively, whose improvement reaches 2.3\% (mAP) and 5.3\% (rank-1). After masking the clothing area, the model can focus on the shoulder contour shape information rather than the texture features of the clothing, which is beneficial to the CC-ReID task. When the foreground image is  employed, highlighting the foreground can further reduce the influence of background interference factors. Note that when the PIE stream is applied, the foreground image and shielding strategy are employed in our experiments.

\begin{table}
\caption{Benefits of the PIE stream on three publicly available, cloth-changing person ReID datasets}
\renewcommand{\arraystretch}{1.4}
\vspace{-1.0em}
\begin{center}
\setlength{\tabcolsep}{1.2mm}{
\begin{tabular}{c|cc|cc|cc}
\hline
\multirow{3}{*}{Methods} &\multicolumn{6}{c} {Datasets} \\
\cline{2-7}
& \multicolumn{2}{c|} {PRCC} &\multicolumn{2}{c|} {Celeb-reID-light} &\multicolumn{2}{c} {VC-Clothes} \\
\cline{2-7}
&mAP &rank-1 &mAP &rank-1 &mAP &rank-1\\
\hline
Baseline                    & 46.4 & 46.3 & 17.1 & 30.2 & 70.6 & 71.2 \\
+PIE [Orig]                    & 47.3 & 46.7 & 22.0 & 36.9 & 73.3 & 72.9 \\
+PIE [Orig w/ Shielding]      & 47.0 & 47.3 & 22.2 & 37.3 & 75.4 & 75.1 \\
\textbf{+PIE [FG w/ Shielding]}   & \textbf{49.3} & \textbf{52.6} & \textbf{22.2} & \textbf{37.3} & \textbf{76.3} & \textbf{75.9} \\
\hline
\end{tabular}}
\end{center}
\vspace{-1.5em}
\label{PIE-Stream-Table}
\end{table}

\begin{table}
\caption{Advantages of the CAD, SAJ, and PIE modules.}
\renewcommand{\arraystretch}{1.4}
\vspace{-1.0em}
\begin{center}
\setlength{\tabcolsep}{1.2mm}{
\begin{tabular}{c|cc|cc|cc}
\hline
\multirow{3}{*}{Methods} &\multicolumn{6}{c} {Datasets} \\
\cline{2-7}
& \multicolumn{2}{c|} {PRCC} &\multicolumn{2}{c|} {Celeb-reID-light} &\multicolumn{2}{c} {VC-Clothes} \\
\cline{2-7}
&mAP &rank-1 &mAP &rank-1 &mAP &rank-1\\
\hline
Baseline         & 46.4 & 46.3 & 17.1 & 30.2 & 70.6 & 71.2 \\
+CAD             & 51.8 & 54.2 & 21.3 & 37.8 & 71.8 & 72.9 \\
+SAJ             & 59.5 & 63.5 & 22.8 & 39.5 & 76.6 & 77.3 \\
+PIE             & 49.3 & 52.6 & 22.2 & 37.3 & 76.3 & 75.9 \\
+CAD+SAJ         & \textbf{60.1} & 63.9 & 22.9 & 39.7 & 77.3 & 78.6 \\
+CAD+SAJ+PIE     & 59.5 & \textbf{65.6} & \textbf{23.4} & \textbf{41.0} & \textbf{77.7} & \textbf{79.0} \\
\hline
\end{tabular}}
\end{center}
\vspace{-1.5em}
\label{collaborative_Stream_Table}
\end{table}

\begin{figure*}
\begin{center}
\includegraphics[width=2.3in,height = 1.64in]{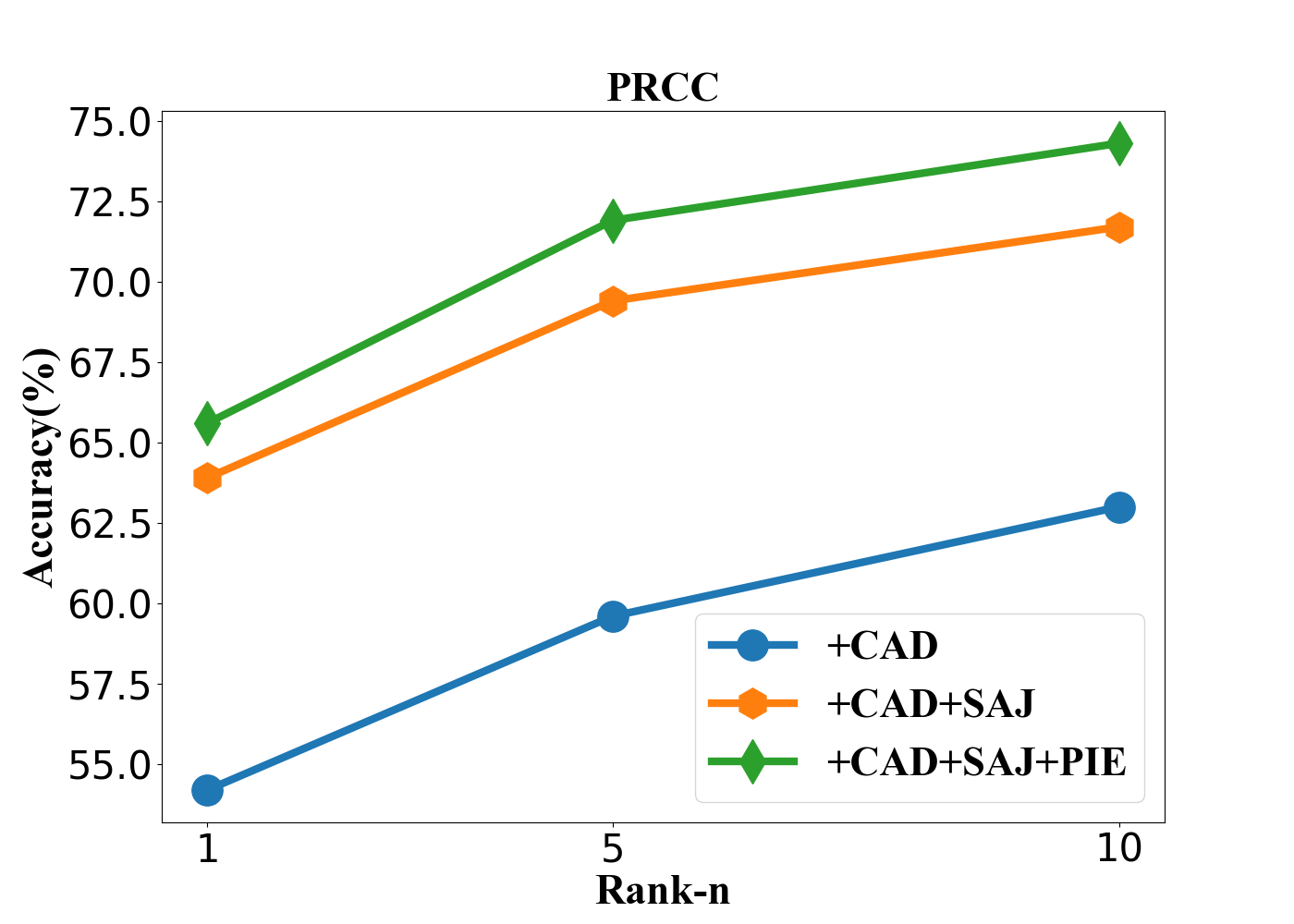}
\hfill
\includegraphics[width=2.3in,height = 1.64in]{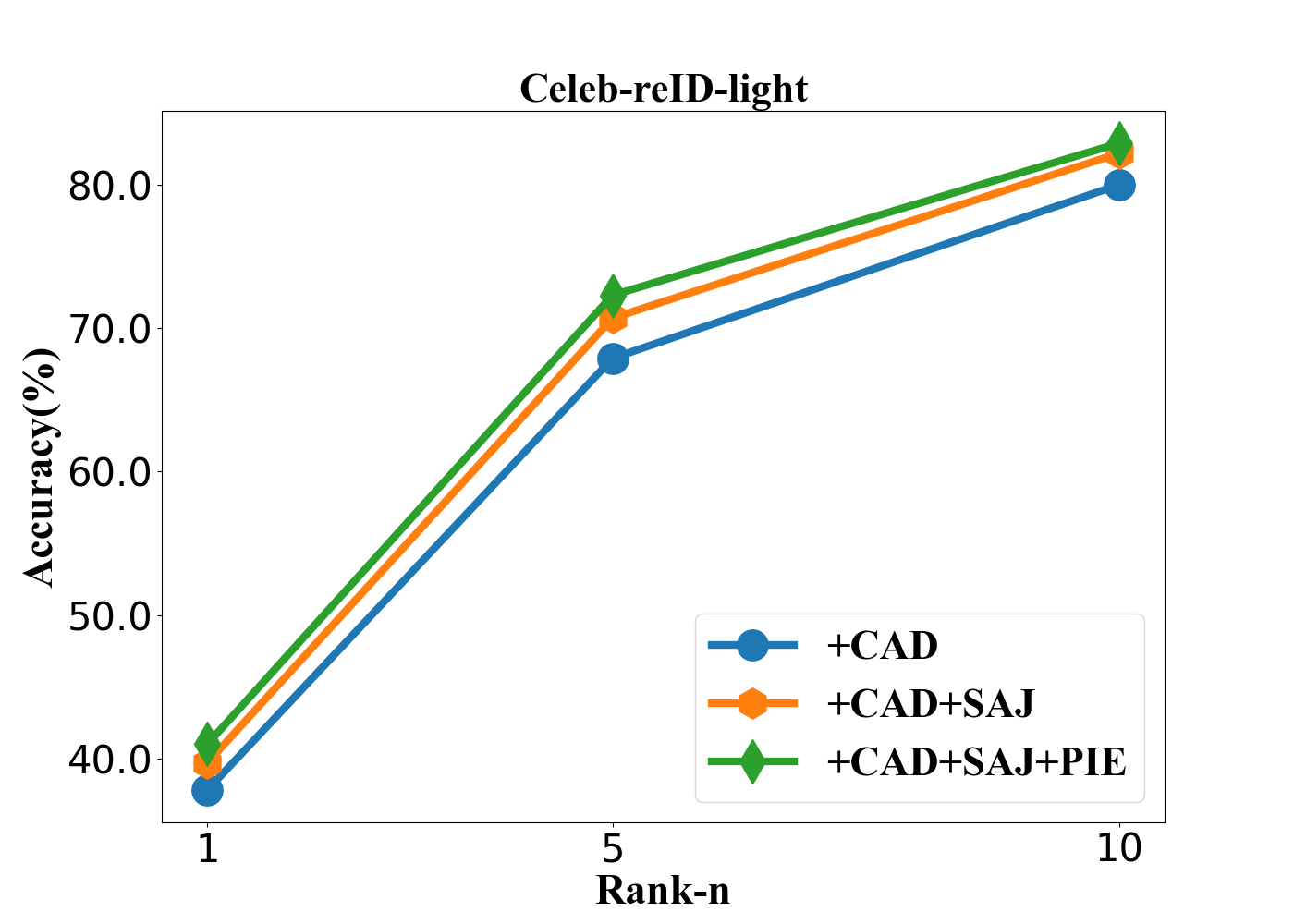}
\hfill
\includegraphics[width=2.3in,height = 1.64in]{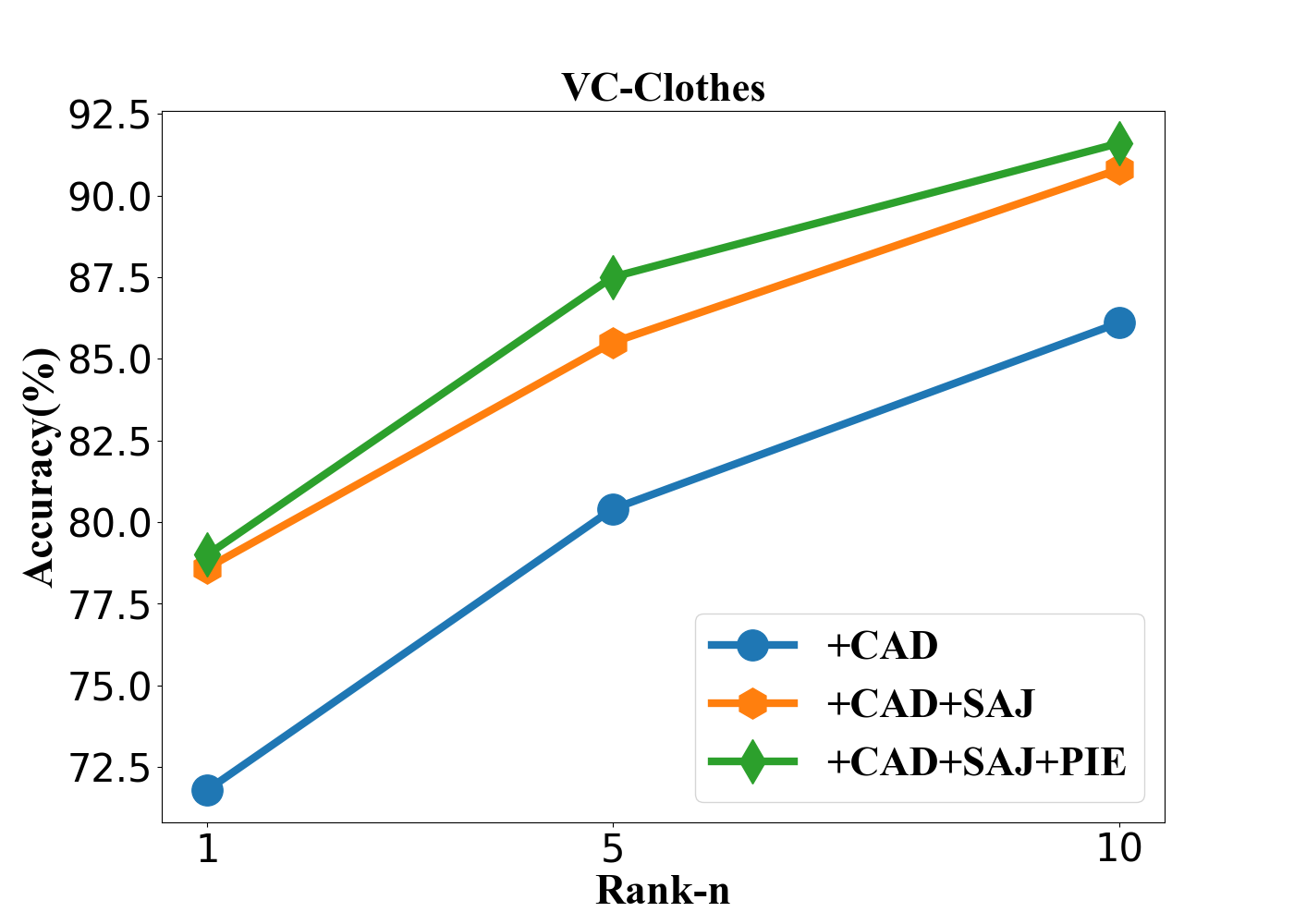}
\caption{Advantages of the PIE stream by using CMC curves on the PRCC, Celeb-reID-light, and VC-Clothes datasets.} \label{collaborative_Stream_Figure}
\end{center}
\vspace{-1.5em}

\end{figure*}

2) To further evaluate the advantages of the PIE stream in enhancing pedestrian identity information, the collaborative learning scheme and CMC curve were employed. In our experiments, the CAD stream, SAJ stream, and PIE stream were embedded in the baseline, and then the collaborative learning scheme, where different streams are jointly optimized in an end-to-end network architecture, was utilized for the CAD stream, SAJ stream, and PIE stream. The results are given in Table \ref{collaborative_Stream_Table} and Figure \ref{collaborative_Stream_Figure}. We observe that when each steam is progressively embedded in the baseline, their combined performance yields a steady boost (except for the mAP of the PRCC dataset). Moreover, these streams are complementary and promote each other. For example, when using the PRCC dataset, the rank-1 accuracy of the baseline, '+CAD', '+CAD+SAJ', '+CAD+SAJ+PIE' are 46.3\%, 54.2\%, 63.9\%, 65.6\%, respectively, and its performance is gradually improved. Moreover, the rank-1 improvement in '+CAD+SAJ+PIE' reaches 19.3\% and 1.7\% compared with the baseline and '+CAD+SAJ'. Similarly, on the Celeb-reID-light dataset, the rank-1 improvement in '+CAD+SAJ+PIE' over the baseline and '+CAD+SAJ' reaches 10.8\% and 1.3\%, respectively. In addition, when the CMC curves are utilized as the metric, we also observe the same results, as shown in Figure \ref{collaborative_Stream_Figure}. Thus, these results establish  the effectiveness of the PIE stream.

\begin{figure}[t]
\begin{center}
\includegraphics[width=3.5in,height = 2.05in]{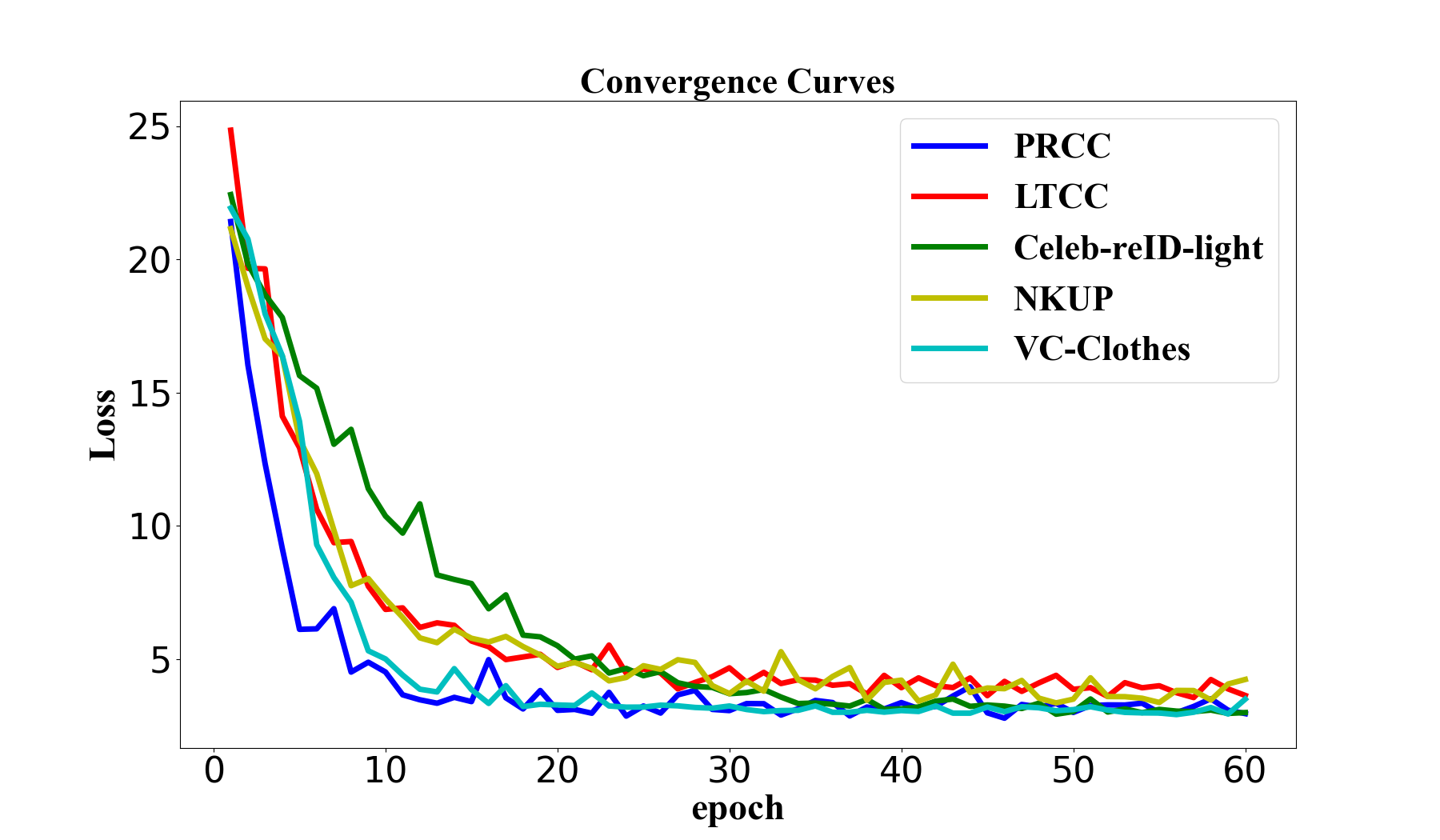}
\caption{Convergence curves of IGCL on the PRCC, LTCC, Celeb-reID-light, NKUP, and VC-Clothes datasets } \label{Convergence_Figure}
\end{center}
\vspace{-0.5em}
\end{figure}

\subsection{Convergence Analysis}
We also evaluate the convergence of the proposed IGCL on the PRCC, LTCC, Celeb-reID-light, NKUP, and VC-Clothes cloth-changing person Re-ID datasets. The corresponding convergence curves are shown in Figure \ref{Convergence_Figure}, where the x-coordinate denotes the number of epochs, and the y-coordinate indicates the loss value. The figure shows that the IGCL method quickly converges regardless of which dataset is employed. In particular,  only 30 to 40 epochs are required for all datasets during the optimization process. Moreover, the convergence curves are very stable regardless of the dataset utilized. This finding further demonstrates the effectiveness of IGCL.


\subsection{Complexity Analysis.}
\begin{table}
\caption{Complexity analysis and the cost of training and testing.  Note that 'Param.' indicates the number of parameters of the model, 'Train time' represents the training time of the model, and 'Test time' is the total 
inference
 time of the model to retrieve all queries.  'm' and 's' denote minutes and seconds, respectively.}
\renewcommand{\arraystretch}{1.2}
\vspace{-1.0em}
\begin{center}
\setlength{\tabcolsep}{1.1mm}
{
\begin{tabular}{c|c|c|c|c|c|c}
\hline
\multirow{2}{*}{Methods} & \multirow{2}{*}{Backbones} & \multicolumn{5}{c}{PRCC}              \\ \cline{3-7}
                         &                            & Param. & Train time & Test time & mAP  & Rank1 \\  \hline     
CAL                      & ResNet                     & 23.52M    & 104m  & 38s      & 55.8 & 55.2  \\
\textbf{IGCL}            & ResNet                     & 85.16M    & 307m  & 40s       & \textbf{57.4} & \textbf{57.8}  \\ \hline
MVSE                     & Densenet                   & 29.56M    & 572m  & 34s      & 52.5 & 47.4  \\
\textbf{IGCL}            & Densenet                   & 51.3M     & 355m  & 43s      & \textbf{59.2} & \textbf{63.1}  \\ \hline
CAL*                     & ViT                        & 86.64M    & 80m   & 27s   & 33.6 & 32.7  \\
MVSE*                    & ViT                        & 102.27M   & 1029m &51s     & 28.8 & 27.3  \\
\textbf{IGCL}            & ViT                        & 233.96M   & 287m  & 39s    & \textbf{63.0} & \textbf{64.4} \\ 
\hline
\end{tabular}
}
\end{center}
\vspace{-1.0em}
\label{model_cost}%
\end{table}

To further assess the performance of different methods, we calculated the number of module parameters, training time, and inference time on the PRCC dataset. All approaches were tested on the same platform with a Tesla A100 GPU, Intel(R) Xeon(R) Gold 6226 CPU @ 2.70 GHz, and 503 GB RAM. We calculated model costs and parameters for IGCL, CAL, and MVSE using ResNet, DenseNet, and Vision Transformer as backbones. Inference time is critical and hence we also evaluated it with different backbones. From the results, we can observe that IGCL has more parameters than CAL or MVSE when utilizing the same backbone, especially for Vision Transformer. Its training time is shorter than that of MVSE's but longer than that of CAL's. The inference time for IGCL is comparable to those for MVSE and CAL. Most importantly, when Vision Transformer is used as the IGCL backbone, its performance is the best among all methods. These results demonstrate the effectiveness of the proposed IGCL with Vision Transformer.

\vspace{1em}

\subsection{Qualitative Visualization}
\begin{figure}
\begin{minipage}{1\linewidth}
\includegraphics[width=3.5in,height = 0.6in]{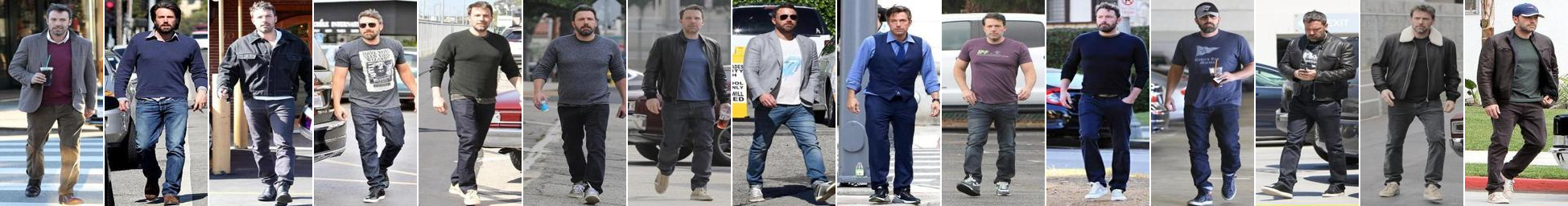}
\centering{(a) 15 images of the same person wearing different clothes} 
\end{minipage}
\\
\begin{minipage}{0.49\linewidth}
\centering{\includegraphics[width=1.6in,height = 1.6in]{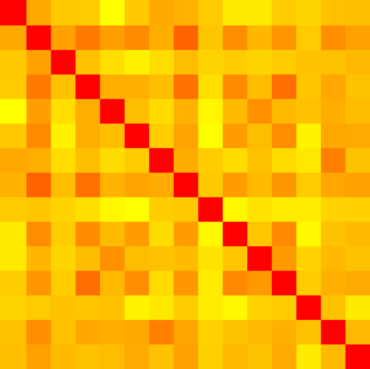}}
\centering{(b) similarity map with the Baseline}
\end{minipage}
\hfill
\begin{minipage}{0.49\linewidth}
\centering{\includegraphics[width=1.6in,height = 1.6in]{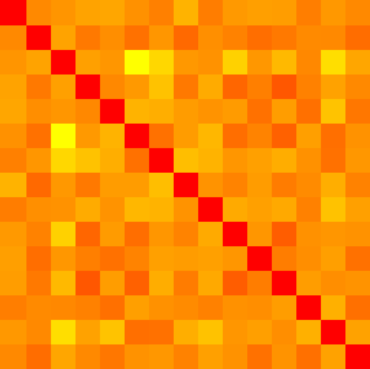}}
\centering{(c) similarity map with the IGCL}
\end{minipage}
\vspace{1em}

\begin{minipage}{1\linewidth}
\includegraphics[width=3.5in,height = 0.6in]{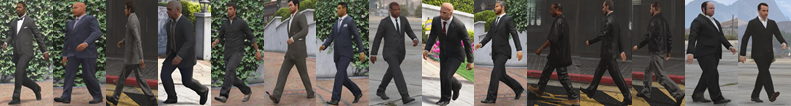}
\centering{(d) 15 images of different people wearing similar clothes}
\end{minipage}
\\
\begin{minipage}{0.49\linewidth}
\centering{\includegraphics[width=1.6in,height = 1.6in]{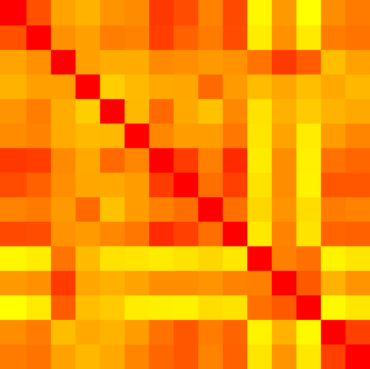}}
\centering{(e) similarity map with the Baseline}
\end{minipage}
\hfill
\begin{minipage}{0.49\linewidth}
\centering{\includegraphics[width=1.6in,height = 1.6in]{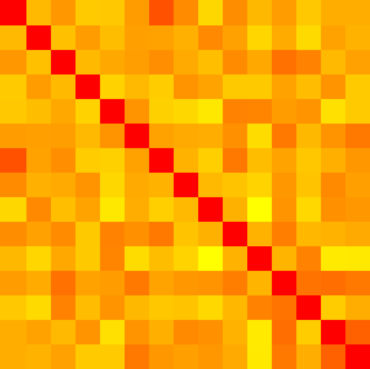}}
\centering{(f) similarity map with the IGCL} 
\end{minipage} 
\caption{Similarity matrices of the baseline and IGCL. Cosine similarity is used to calculate the distance between any two images. The color of each square indicates the similarity degree between these two images indicated by the horizontal and vertical coordinates. The red and yellow colors represent the most similar pairs and least similar pairs, respectively.} \label{Person_Similarity_matrices}
\vspace{-0.5em}
\end{figure}

\begin{figure*}
\begin{center}
\includegraphics[width=7in,height = 4.872in]{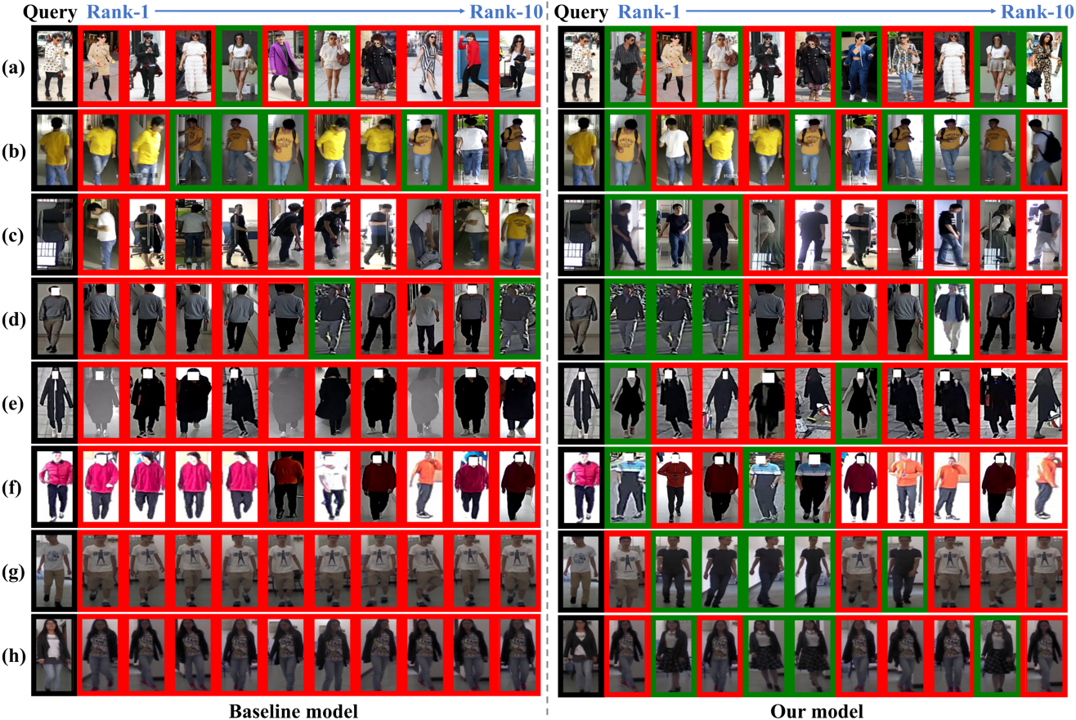}
\caption{Top-10 ranking results of our IGCL and the baseline where different queries with
different cases, such as background clutter (a), side and back images (b, g-h), partial occlusion (c), and covered face information (d-f), are utilized. Green boxes indicate correct results and red boxes represent incorrect results.} \label{Top_10_ranking_results}
\end{center}
\vspace{-1.5em}
\end{figure*}

\begin{figure*}
\begin{center}
\includegraphics[width=7in,height = 2.0in]{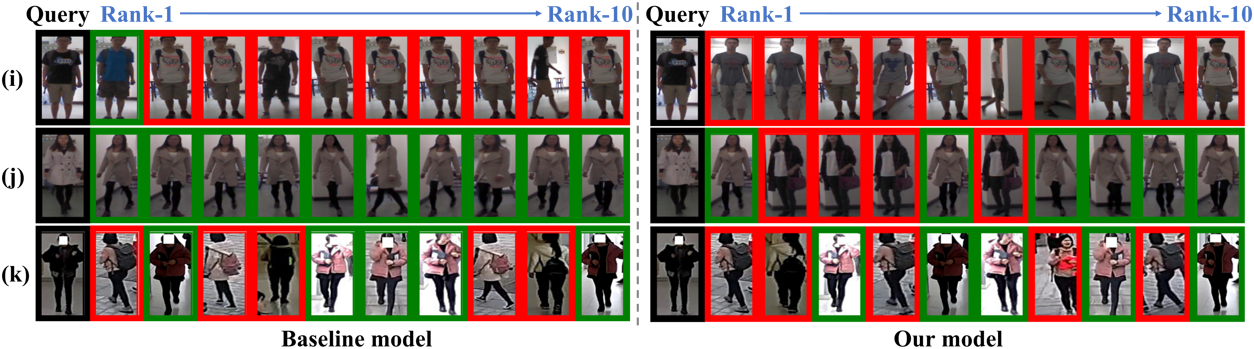}
\caption{Several cases are presented when IGCL fails or lags behind} \label{Failed_Case}
\end{center}
\vspace{-2.5em}
\end{figure*}

To further prove the effectiveness of the IGCL, we visualized some of the experimental results while considering three aspects: 1) visualization of the attention maps, 2) visualization of the similarity map, and 3) qualitative visualization of the retrieval results. In what follows, we separately discuss these three aspects; the results are given in Figures \ref{attention_maps}, \ref{Person_Similarity_matrices}, \ref{Top_10_ranking_results} and \ref{Failed_Case}. We make the following observations:

\begin{figure}
\begin{center}
\includegraphics[width=3.5in,height = 2.594in]{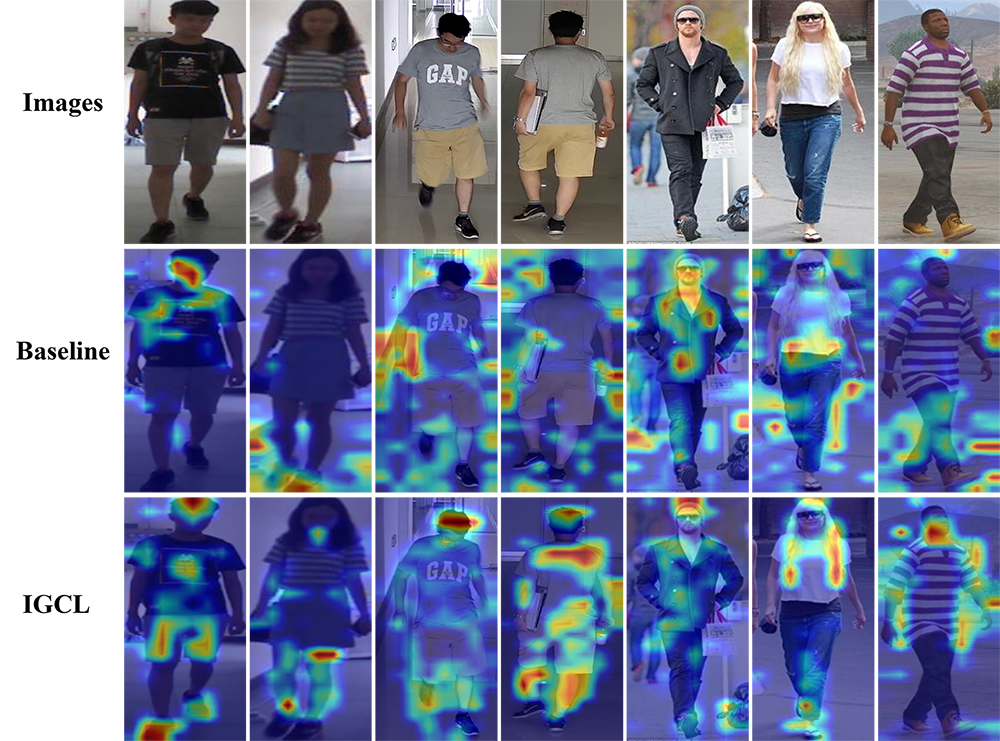}
\caption{Visualization of attention maps. The first row, second row, and third row indicate the original images, attention maps of the baseline, and attention maps of IGCL, respectively. The brighter the pixels, the more attention is paid by the model, and the identity of each column belongs to the same person.} \label{attention_maps}
\end{center}
\vspace{-1.5em}
\end{figure}

1) To better understand the working principle of different modules and to further illustrate which cues are more focused, we used the Grad-CAM \cite{selvaraju2017grad} method to visualize and display the intermediate activation feature maps of the baseline and IGCL in Figure \ref{attention_maps}. We observe that the baseline feature maps mainly focus on the global context information and clothing texture information. Much interference information (such as background information) is introduced into the feature extraction, whose feature discrimination, clothing-irrelevance, and generalization need to be further improved. In contrast, the activation feature maps of IGCL pay more attention to the overall human body structure (different parts of the human) and the local head, neck, shoulder, and shoe information (It is worth noting that shoes change less frequently than clothes, and everyone's shoes have a certain identity bias and thus, it is robust to cloth-changing) while paying minimal attention to places containing more interference factors such as texture information of clothing, background, etc. Therefore, these experiments further demonstrate the effectiveness and advantages of our proposed method.

2) To intuitively illustrate the effectiveness of the IGCL from another perspective, the feature similarity between two different images was calculated. We chose the data from the Celeb-reID-light dataset, which has the highest complexity of cloth-changing. In our experiments, we selected 15 images of the same person wearing different clothes, and then the baseline module was used to extract the features for each image. Moreover, the cosine similarity between any two images was calculated by the corresponding extracted features. The reason why cosine similarity was utilized is that the normalized similarity matrices need to be obtained; thus, cosine similarity was used to measure the matching scores between different pedestrian image feature vectors extracted by the baseline and IGCL models whose values vary from 0 to 1. We repeated the above operation in pairs for all 15 images and visualized their similarities to obtain a similarity matrix of $15 \times15$. The IGCL module was also used to extract the feature representations for all 15 images, and we also calculated the cosine similarity and similarity matrix between them by these new features. The results are given in Figure \ref{Person_Similarity_matrices}.

From the results, we can observe that when only the baseline is used, the extracted features are correlated with interference factors such as background and clothing. This results in low similarity scores for the same person wearing different clothes. However, when IGCL is employed, the baseline is learned collaboratively with the CAD stream, SAJ stream, and PIE stream. This effectively mines more clothing-independent cues and robust identity information. This in turn leads to more discriminative and clothing-irrelevant features, resulting in higher similarity scores for the same person wearing different clothes. The lighter color in the similarity matrix indicates a low correlation between elements, while darker colors indicate higher similarity. The pedestrian image in Figure \ref{Person_Similarity_matrices}(a) shows that the background and clothing are the areas with the largest difference, suggesting that the model may pay too much attention to them, leading to low similarity. Conversely, stable body and face areas reflect a high similarity, as seen in the darker color in Figure \ref{Person_Similarity_matrices} (c). These experiments further demonstrate the effectiveness of IGCL in focusing on stable areas for obtaining improved similarity scores.

We also selected 15 images of different people wearing similar clothes from the VC-Clothes dataset, as shown in Figure \ref{Person_Similarity_matrices}(d). When only the baseline was used, the similarity feature map showed greater similarity between different people wearing similar clothes, as shown in Figure \ref{Person_Similarity_matrices}(e). This indicates that the extracted features may contain more clothing information. However, when IGCL is employed, the similarity feature map tends to become lighter, indicating a decrease in similarity between different people wearing similar clothes. This suggests that IGCL can suppress similar clothing information to some extent and focus on more nonclothing discrimination features.

3) To further demonstrate the effectiveness of the IGCL method, the visualization retrieval results of IGCL and the baseline are shown in Figures \ref{Top_10_ranking_results} and \ref{Failed_Case}, where each row is a retrieval example. We observe that although the baseline is trained by the training samples with different clothing identities, additional guidance or regularization is not employed. Thus, the appearance features are inevitably disturbed by clothing information, especially by clothing texture and the background. For example, in row (f) in Figure \ref{Top_10_ranking_results}, the query sample is a man wearing a red top and black pants. The corresponding unmatched samples obtained by the baseline have highly similar clothing color patterns. In contrast, the appearance features extracted by IGCL demonstrate a stronger correlation with identity across both low-level visual attributes and high-level semantics, surpassing the appearance features obtained by the baseline method. As a result, the proposed method holds the potential to enhance identity invariance in specific scenarios and aims to accurately discern and eliminate incongruent samples that exhibit similar clothing textures.  We reach similar conclusions from other rows, such as Figure \ref{Top_10_ranking_results} (b), (d), and (e). In addition, when the query image background is very cluttered (Figure \ref{Top_10_ranking_results}(a)) or partial occlusion (Figure \ref{Top_10_ranking_results}(c)) occurs, the performance of the baseline is often unsatisfactory. In this case, IGCL still effectively retrieves 3-5 correct results, which are also at the top of the returned results. Moreover, when the provided query image is a backside image in Figure \ref{Top_10_ranking_results}(b) or the face is occluded in Figure \ref{Top_10_ranking_results}(e), IGCL can still correctly identify the person, but the average result significantly decreases, where only one or two correct results are returned. These experimental results show that it is very challenging to perform a cloth-changing person-ReID task when the provided images largely lack visual semantics. Moreover, the visualization results also prove that the proposed IGCL can effectively explore pedestrian identity information and that collaborative learning is very helpful for overcoming the challenge of the cloth-changing person ReID task.

However, since clothing information plays a significant role in the overall contextual data, while IGCL can reduce interference from clothing-related details, it cannot completely eliminate them. Therefore, during instances of clothing changes, especially when individuals assume different poses, there is still a possibility of incorrect matches. Despite IGCL's superior performance compared to the baseline, it is essential to acknowledge the potential for occasional erroneous matches.  We also present some failed and lagging cases of IGCL in  Figure \ref{Failed_Case}.  While IGCL weakens the sensitivity of clothing information, it still relies on posture, hairstyle, body shape, accessories, and other information to varying degrees. This can lead to matching failures. For example, in Figure \ref{Failed_Case}(j), the human body of the query image is similar to the rank-1 image in body shape; thus, it is easy for the baseline to find it, but IGCL pays more attention to the hairstyle, head, feet, etc.; thus, IGCL can find a similar image with the same hairstyle and feet. Similar cases arise for other query images. This reflects the need to explore how to effectively balance different pieces of relevant information in IGCL.

\section{Conclusion}
In this work, we propose a novel method named IGCL to exploit robust and informative pedestrian representations for the cloth-changing person ReID task. The key idea is to shield clues related to the appearance of clothes and focus only on human semantics and identity information. A CAD stream is designed to reduce the interference caused by clothing information, and a SAJ stream is constructed to highlight human semantic information and simulate different poses of the same identity. A PIE stream is built to enhance the identity importance, where only the information of the head and shoulders are effectively employed. Most importantly, all these streams are collaboratively learned in an end-to-end unified framework. The results of extensive experiments conducted on six cloth-changing person ReID datasets validate the effectiveness of our proposed IGCL framework. In particular, IGCL outperforms SOTA cloth-changing person ReID methods in terms of the accuracy of mAP and rank-1 on multiple datasets. Moreover, more discriminative, robust, and clothing-irrelevant features are extracted to describe pedestrians with different clothes. In addition, our ablation study proves that human semantic and identity information and collaborative learning are very helpful for solving the cloth-changing person ReID task. In the future, we intend to focus on real-person ReID scenarios and design a large-scale person ReID module that can be effectively applied to different ReID tasks, e.g., holistic person ReID, partial person ReID, occluded person ReID, and cloth-changing person ReID.

\section{Acknowledge}
This work was supported in part by the National Natural Science Foundation of China (No.62372325, No.61872270), Young creative team in universities of Shandong Province (No.2020KJN012), Jinan 20 projects in universities (No.2020GXRC040). Shandong project towards the integration of education and industry (No.2022PYI001, No. 2022PY009, No.2022JBZ01-03). CCF- Baidu Open Fund under Grant CCF-BAIDU OF2022008.

 \normalem
 \bibliographystyle{IEEEtran}
 \bibliography{mybib}

%

\begin{IEEEbiography}[{\includegraphics[width=1in,clip,keepaspectratio]{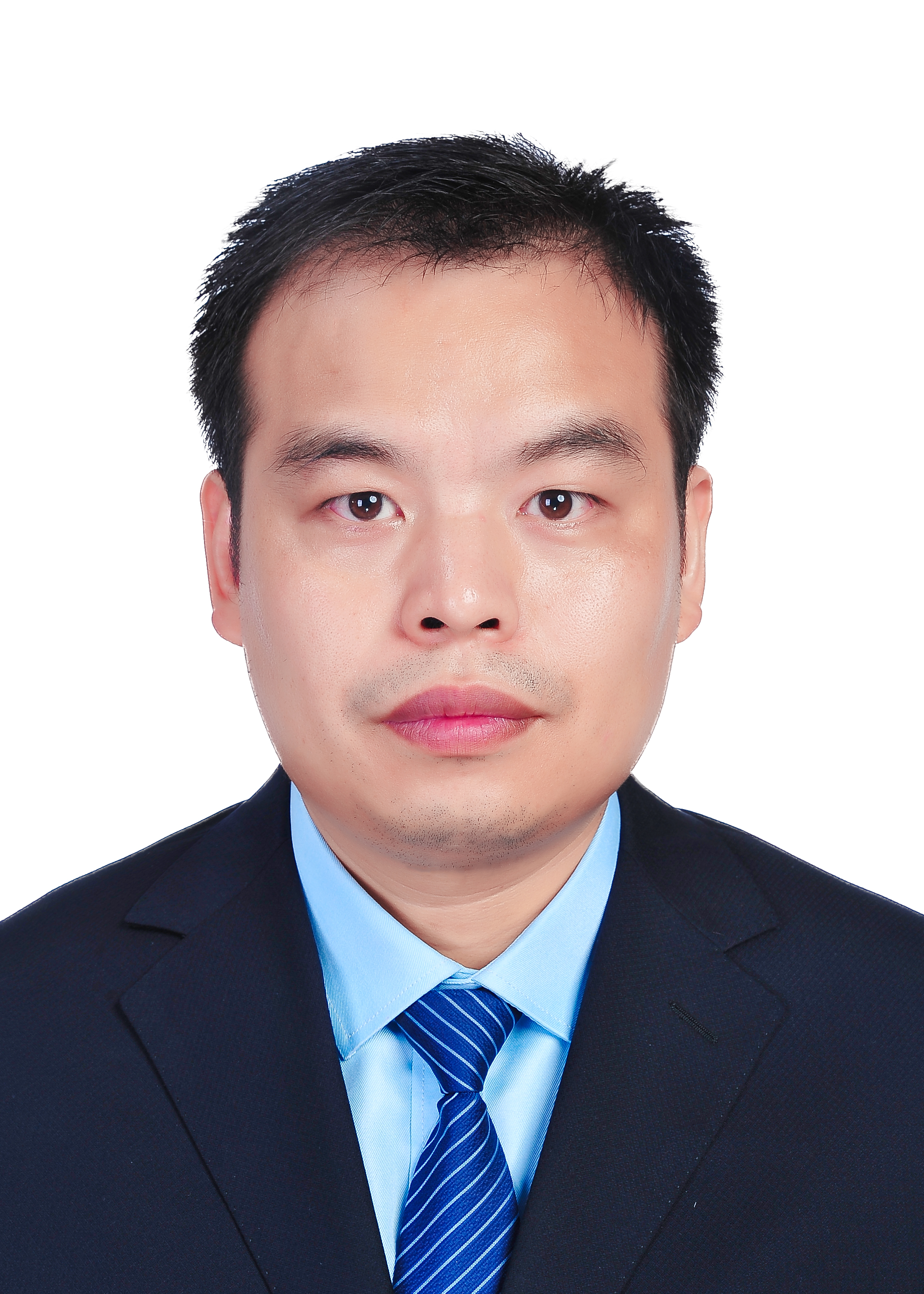}}]{Zan Gao} received his Ph.D degree from Beijing University of Posts and Telecommunications in 2011. He is currently a full Professor with the Shandong Artificial Intelligence Institute, Qilu University of Technology (Shandong Academy of Sciences). From Sep. 2009 to Sep. 2010, he worded in the School of Computer Science, Carnegie Mellon University, USA. From July 2016 to Jan 2017, he worked in the School of Computing of National University of Singapore. His research interests include artificial intelligence, multimedia analysis and retrieval, and machine learning.  He has authored over 100 scientific papers in international conferences and journals including TPAMI, TIP, TNNLS, TMM, TCYBE, TOMM, CVPR, ACM MM, WWW, SIGIR and AAAI, Neural Networks, and Internet of Things.
\end{IEEEbiography}

\begin{IEEEbiography}[{\includegraphics[width=1in,clip,keepaspectratio]{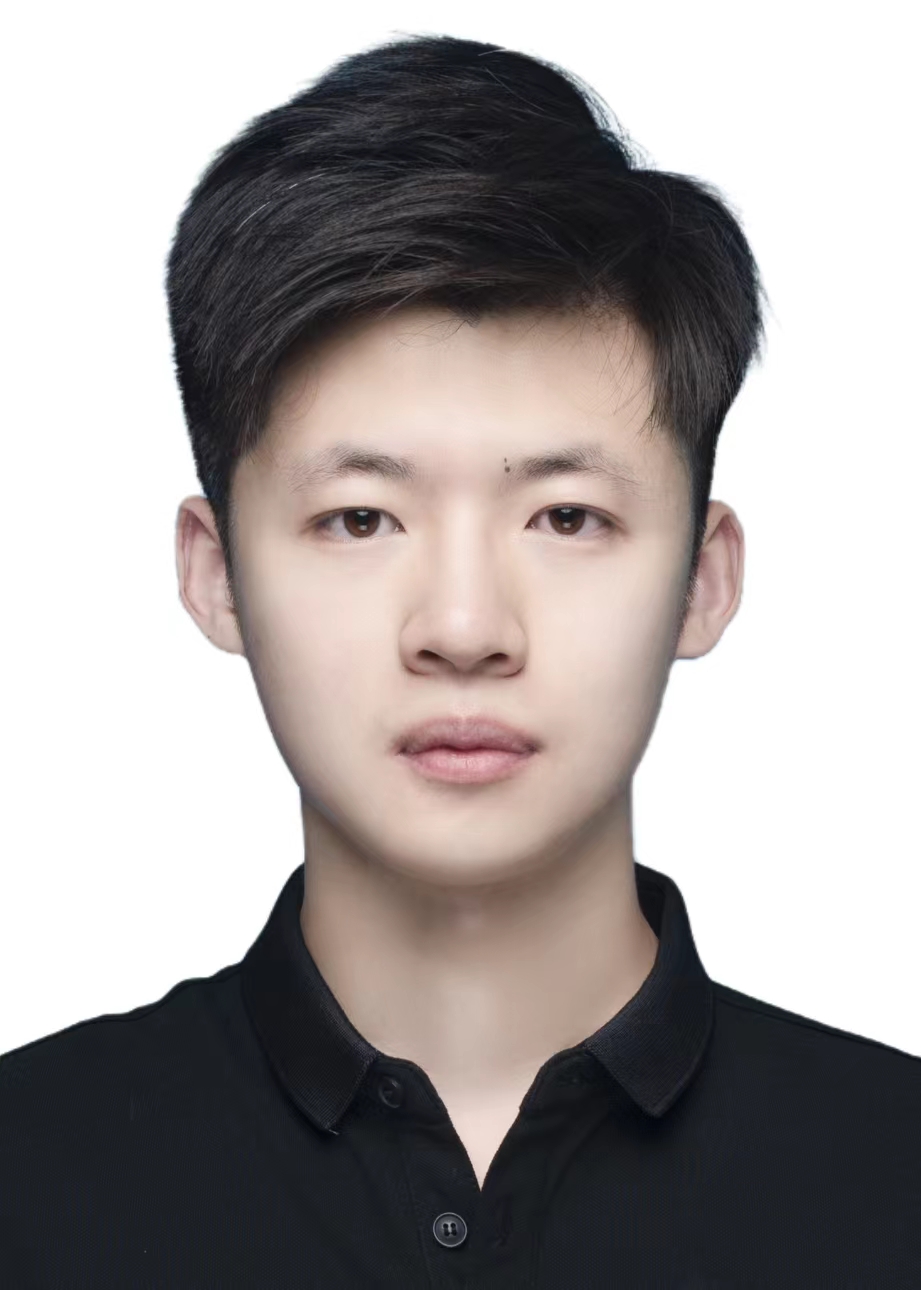}}]{Shengxun Wei} is pursuing his master degree in the Shandong Artificial Intelligence Institute, Qilu University of Technology (Shandong Academy of Sciences). His research interests include artificial intelligence, multimedia analysis and retrieval, computer vision and machine learning.
\end{IEEEbiography}

\begin{IEEEbiography}[{\includegraphics[width=1in,height=1.25in]{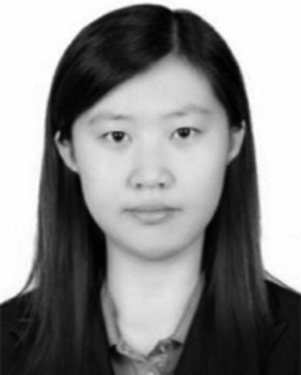}}]{Weili Guan} is now a Ph.D. student with the Faculty of Information Technology, Monash University Clayton Campus, Australia. Her research interests are multimedia computing and information retrieval. She received her bachelor degree from Huaqiao University in 2009. She then obtained her graduate diploma and master degree from National University of Singapore in 2011 and 2014 respectively. After that, she joined Hewlett Packard enterprise Singapore as software engineer and worked there for around five years. 
\end{IEEEbiography}

\begin{IEEEbiography}[{\includegraphics[width=1in,clip,keepaspectratio]{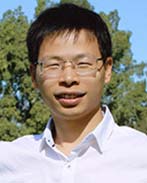}}]{Lei Zhu} received the B.S. degree from the Wuhan University of Technology, Wuhan, China, in 2009, and the Ph.D. degree from  the Huazhong University of Science and Technology, Wuhan, in 2015. He was a Research Fellow with The University of Queensland, St Lucia, QLD, Australia, from 2016 to 2017, and Singapore Management University, Singapore, from 2015 to 2016. He is currently a Full Professor with the School of Electronic and Information Engineering, Tongji University, Shanghai, China. His research interests are in the area of large-scale multimedia content analysis and retrieval.
\end{IEEEbiography}


\begin{IEEEbiography}[{\includegraphics[width=1in,clip,keepaspectratio]{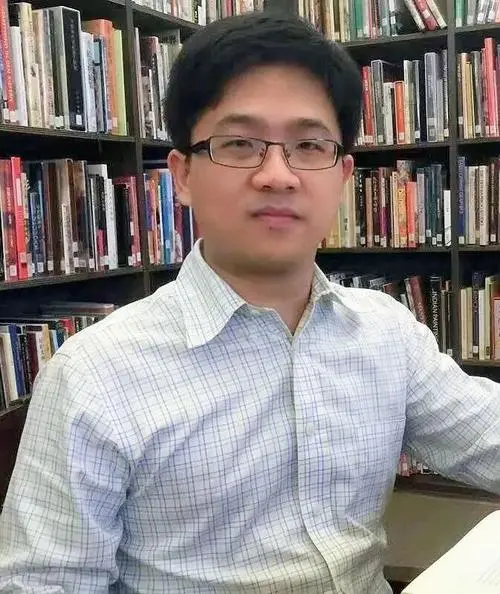}}]{Meng Wang} received the BE and PhD degrees in the Special Class for the Gifted Young and the Department of Electronic Engineering and Information Science from the University of Science and Technology of China (USTC), Hefei, China, respectively. He is a professor at the Hefei University of Technology, China. His current research interests include multimedia content analysis, search, mining, recommendation, and large-scale computing. He received the best paper awards successively from the 17th and 18th ACM International Conference on Multimedia, the best paper award from the 16th International Multimedia Modeling Conference, the best paper award from the 4th International Conference on Internet Multimedia Computing and Service, and the best demo award from the 20th ACM International Conference on Multimedia.
\end{IEEEbiography} 

\begin{IEEEbiography}[{\includegraphics[width=1in,clip,keepaspectratio]{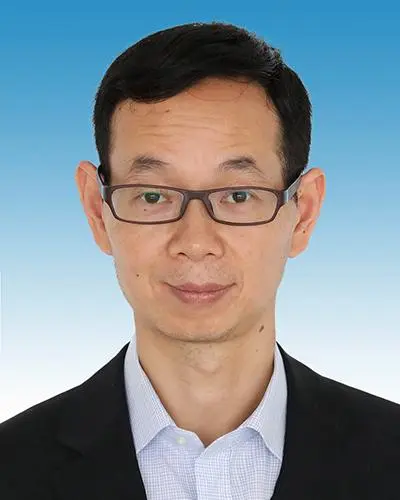}}]{Shengyong Chen} received the Ph.D. degree in computer vision from the City University of Hong Kong, Hong Kong, in 2003. He was with the University of Hamburg from 2006 to 2007. He is currently a Professor at the Tianjin University of Technology and also with the Zhejiang University of Technology, China. He received the Fellowship from the Alexander von Humboldt Foundation of Germany. He has authored over 100 scientific papers in international journals. His research interests include computer vision, robotics, and image analysis. He is a fellow of the IET and a Senior Member of the CCF. He received the National Outstanding Youth Foundation Award of China in 2013.
\end{IEEEbiography}




\end{document}